\documentclass{article}
\PassOptionsToPackage{numbers, compress}{natbib}

\usepackage[preprint]{neurips_2026}

\usepackage[utf8]{inputenc} 
\usepackage[T1]{fontenc}    
\usepackage{hyperref}       
\usepackage{url}            
\usepackage{booktabs}       
\usepackage{graphicx}       
\usepackage{subcaption}     
\usepackage{enumitem}       
\usepackage{amsmath}        
\usepackage{amssymb}        
\usepackage{amsthm}         
\usepackage{xspace}         
\usepackage{amsfonts}       
\usepackage{nicefrac}       
\usepackage{microtype}      
\usepackage{xcolor}         
\usepackage{caption}
\usepackage{multirow}
\usepackage{wrapfig}
\usepackage{longtable}

\hypersetup{
  colorlinks=true,
  linkcolor=blue,
  citecolor=blue,
  urlcolor=blue
}
\title{Geometry-Lite: Interpretable Safety Probing via Layer-Wise Margin Geometry}

\author{
  \AND
  Woo Seob Sim\thanks{Department Biomedical Systems Informatics Yonsei University College of Medicine, Seoul, Republic of Korea}\\
  Yonsei University\\
  \texttt{dogu86@yuhs.ac}\\
  \And
  Yu Rang Park\footnotemark[1] \hspace{1.7mm} \thanks{Corresponding Author. yurangpark@yuhs.ac}\\
  Yonsei University\\
  \texttt{yurangpark@yuhs.ac}
\\  
}

\begin{document}

\maketitle

\begin{abstract}
Prompt-level safety probes for large language models use hidden-state
representations to separate safe from unsafe prompts, but strong average
detection performance does not explain the geometry of this separation.
In particular, it remains unclear how safety evidence is formed across
layers, which aspects of that layer-wise geometry support
low-false-positive decisions, and which geometric biases remain stable
under benchmark shift. We study this as an
empirical decomposition problem and introduce Geometry-Lite, a compact
prompt-level probe that maps each layer's final prompt-token
representation to signed margins under centroid, local-neighborhood, and
supervised linear-boundary readouts, then summarizes the resulting margin
profiles by boundary position, layer-to-layer change, and coarse shape.
Across nine instruction-tuned backbones ($1.2$B--$70$B) and seven safety
benchmarks, Geometry-Lite improves over single-layer probes while
remaining close to raw multi-layer score stacking, making it a useful
instrument for analyzing the multi-layer safety signal. The decomposition
shows that safety evidence is expressed primarily through persistent
boundary-position geometry: final or extremal margins and unsafe-side
layer occupancy dominate aggregate detection performance. In contrast,
finite-difference drift and structural summaries add little to pooled
AUROC, although drift can provide small recall-oriented corrections under
shifted low-FPR thresholds. Under benchmark shift, optimized linear
boundaries are sharp on the training mixture, whereas class-conditional
mean geometry retains separation more reliably on a predefined hard
held-out subset. Overall, prompt-level safety evidence is not primarily a
layer-to-layer motion signal, but a persistent layer-wise margin geometry
whose useful components and readout-level biases become visible in
decision-critical regimes.
\end{abstract}

\section{Introduction}
\label{sec:intro}

As large language models are deployed at scale, prompt-level harmfulness 
detection has become a practical component of safety guardrails 
\citep{inan2023llamaguard,markov2023holistic}. Hidden-state probes are
attractive in this setting because they can classify a prompt before
generation while also exposing where safety-relevant information appears
inside the model. A common realization trains a lightweight linear
classifier on a single hidden-state snapshot, typically the final
prompt-token representation at a chosen middle or last layer
\citep{marks2023geometry,arditi2024refusal}. Such snapshot probes are
cheap to fit, strong on standard benchmarks, and easy to inspect, which
has made them a standard baseline for prompt-level safety classification.

A snapshot probe operationalizes the assumption that a single well-chosen layer provides a sufficient summary of the safety-relevant signal. Two observations make this assumption worth reexamining. First, aggregate performance on standard benchmarks can obscure performance on less obvious or ambiguous prompts. In deployment, these borderline cases are especially important because they determine behavior at the low false-positive operating points used by safety systems \citep{rottger2024xstest, shaikh2023second}. Second, out-of-distribution (OOD) generalization is a long-standing concern in machine learning, since models that perform well on benchmark splits can rely on shortcut features that fail to transfer \citep{geirhos2020shortcut, recht2019imagenet, koh2021wilds}. Recent work documents that activation-based safety probes inherit this difficulty, with substantial degradation under cross-dataset evaluation \citep{fomin2026benchmarks}. Together, these observations motivate a more fine-grained question:
not only whether multiple layers help, but what form of multi-layer
evidence remains useful in low-FPR and shifted regimes.

Several lines of work motivate looking beyond a single hidden-state snapshot. Tuned-lens and interpretability analyses view transformer inference as a layer-wise refinement or prediction-construction process \citep{geva2022transformer, belrose2023tuned}, representation-engineering methods extract task-relevant directions from hidden states \citep{zou2023representation}, and recent trajectory methods model layer-wise displacements rather than static activations \citep{damirchi2026truth}. In safety specifically, prior work has identified refusal or safety directions in internal activations and has localized safety-relevant behavior to selected middle or late layers \citep{arditi2024refusal, li2025safety}. These results support the use of multi-layer evidence, but they do not
determine what kind of layer-wise signal is being used, or which part
survives low-FPR thresholds and benchmark shift. If raw per-layer score stacking is sufficient, the signal is distributed across several informative depths. If displacement models add signal beyond static reads, layer-to-layer motion carries information that per-layer position does not. If boundary-relative summaries are sufficient, the signal is better understood as the position and persistence of a margin profile relative to safety boundaries. Distinguishing these alternatives matters because they can yield similar
aggregate performance while implying different robustness, complexity,
and interpretability tradeoffs; under shift, even the geometry used to
define the layer-wise margin may matter.

To make these alternatives empirically separable, we introduce
\textbf{Geometry-Lite}, a prompt-level safety probe that decomposes
multi-layer signal along labeled geometric axes. For each transformer
layer, we reduce the final prompt-token representation to a scalar margin
under three geometric readouts: class-centroid distance,
$k$-nearest-neighbor neighborhood structure, and a supervised linear
safety boundary. This yields three scalar margin profiles indexed by
depth. We summarize each profile along three named axes,
\emph{margin level}, \emph{layer-to-layer change}, and
\emph{structural shape}, and classify the resulting 39-feature
representation by L2-regularized logistic regression.

We evaluate Geometry-Lite together with raw multi-layer score-stacking
variants and trajectory baselines, treating these methods as controlled
contrasts between different explanations of the multi-layer safety
signal. Across nine instruction-tuned backbones from the Llama, Gemma,
and Qwen families ($1.2$B--$70$B) and seven safety benchmarks,
Geometry-Lite remains competitive with the strongest raw-stacking
variant while exposing which components of the layer-wise signal are
responsible for performance. This lets us separate two questions that
raw aggregate metrics conflate: which summaries of a margin profile carry
the useful signal, and which margin geometry remains stable under
benchmark shift.

Our contributions are as follows.
\begin{itemize}[leftmargin=1.5em,itemsep=2pt,topsep=2pt]
    \item \textbf{Geometry-Lite.}
    We introduce a compact prompt-level probe that converts layer-wise
    hidden states into signed margin profiles under centroid,
    local-neighborhood, and supervised linear-boundary readouts, then
    summarizes them along named axes for boundary position,
    layer-to-layer change, and profile shape.

    \item \textbf{A decomposition of multi-layer safety evidence.}
    We use Geometry-Lite together with raw score-stacking and trajectory
    baselines to separate three explanations of multi-layer signal:
    aggregation across informative layers, boundary-relative margin
    position, and explicit layer-to-layer motion.

    \item \textbf{Findings across low-FPR and shifted regimes.}
    Across nine backbones and seven benchmarks, boundary-position
    summaries explain most aggregate performance. Drift is sparse but can
    provide small recall corrections under shifted low-FPR thresholds.
    Under benchmark shift, optimized linear boundaries are sharper on the
    training mixture, whereas class-mean geometry retains separation more
    reliably on the hardest held-out benchmarks.
\end{itemize}
\section{Related Work}
\label{sec:related}

\paragraph{Hidden-state probes for safety.}
A growing body of work trains linear or shallow classifiers on transformer hidden states to detect harmful prompts or safety attributes. \citet{arditi2024refusal} identify a single refusal-mediating direction obtained from a difference of class means, and \citet{zou2023representation} introduce representation engineering for extracting task-specific reading directions across layers. The same difference-in-means (DIM) construction has been studied as a probing direction in its own right \citep{marks2023geometry}, and we include a multi-layer extension (MultiLayer-DIM) as a baseline. \citet{li2025safety} further localize safety behavior to a small set of contiguous middle layers. These works motivate two of our readout families: class-mean directions,
as in difference-in-means and representation-engineering probes, and
optimized linear boundaries, as in standard hidden-state probing.

\paragraph{Layer-wise and trajectory-based views.}
A parallel line treats transformer inference as a layer-wise process that progressively constructs predictions \citep{geva2022transformer, belrose2023tuned}. Concurrent to our work, \citet{damirchi2026truth} propose Truth as a Trajectory (TaT), modeling layer-wise displacements $\delta_\ell = h_\ell - h_{\ell-1}$ with an LSTM and evaluating across reasoning, QA, and toxicity benchmarks. We include a matched final-token adaptation of their classifier as the Truth-as-a-Trajectory displacement LSTM (TaT-Disp-LSTM) baseline. We use TaT-Disp-LSTM as a matched displacement-based baseline, while
Geometry-Lite summarizes the layer-wise signal along labeled geometric
axes that can be ablated independently. This contrasts a learned
sequence model, which folds position and motion into a single hidden
state, with named margin-profile summaries.

\paragraph{Cross-benchmark evaluation.}
Out-of-Distribution (OOD) generalization is a long-standing concern in machine learning, with models often relying on shortcut features that fail to transfer \citep{geirhos2020shortcut, koh2021wilds}. Within safety probing, \citet{fomin2026benchmarks} adopt a leave-one-dataset-out (LODO) protocol that holds out an entire safety dataset and reports substantial cross-dataset degradation. Our leave-one-benchmark-out (LOBO) protocol operates at a finer granularity within a single safety domain, holding out one of seven safety benchmarks and training on the remaining six. The two protocols are complementary, with LODO measuring transfer across datasets that may differ in domain, and LOBO measuring transfer across benchmarks that share the prompt-level safety task but differ in coverage and harm taxonomy.

\section{Method}
\label{sec:method}

\subsection{Problem definition}
\label{sec:method:problem}

Given a causal chat language model $f$ and a user prompt $x$, we apply
the model's chat template and run a forward pass over the prompt tokens
only. For each layer $\ell$, let $h^{(\ell)}(x)\in\mathbb{R}^{D}$ be the
hidden state at the final prompt-token position, before any response
token is generated. We collect
\begin{equation}
H(x)=\big[h^{(1)}(x),h^{(2)}(x),\ldots,h^{(L)}(x)\big]
\in\mathbb{R}^{L\times D}.
\end{equation}
Given labeled prompts $\{(x_i,y_i)\}_{i=1}^{N}$ with $y_i\in\{0,1\}$
and $y=1$ denoting unsafe, our goal is a prompt-level classifier
$g:\mathbb{R}^{L\times D}\to[0,1]$ estimating $p(y=1\mid x)$ from
pre-generation hidden states.

\subsection{From hidden states to boundary-relative margins}
\label{sec:method:geometries}

For each layer $\ell$, we reduce the $D$-dimensional hidden state to a
scalar \emph{margin} using a reference geometry fit on the training
split. We use three complementary geometries to span a small set of
readout biases: class-conditional mean geometry, which underlies
difference-in-means and representation-engineering probes
\citep{marks2023geometry,arditi2024refusal,zou2023representation};
local neighborhood structure; and an optimized supervised boundary,
corresponding to standard linear hidden-state probes
\citep{inan2023llamaguard,li2025safety}.

\paragraph{Centroid distance.} Let $\mu^+_\ell$ and $\mu^-_\ell$ denote the training-set class means at layer $\ell$ for the safe ($y=0$) and unsafe ($y=1$) classes, respectively. The centroid margin is the 
difference of distances to the two class means:
\begin{equation}
m^{\text{cent}}_\ell(h) = \|h - \mu^-_\ell\|_2 - \|h - \mu^+_\ell\|_2.
\end{equation}
A safe-leaning representation lies closer to $\mu^+_\ell$ than to 
$\mu^-_\ell$, yielding positive margin.

\paragraph{$k$-NN local-neighborhood.} For a hidden vector $h$, let $\mathcal{N}_k^+(h)$ 
and $\mathcal{N}_k^-(h)$ denote its $k$ nearest neighbors under cosine 
distance in the safe and unsafe layer-$\ell$ training pools, respectively. 
Define $\bar{c}^{c}_k(h) = \frac{1}{k} \sum_{h' \in \mathcal{N}_k^{c}(h)}
d_{\cos}(h,h')$ for $c \in \{+,-\}$. The $k$-NN margin is the 
difference of mean cosine distances:
\begin{equation}
m^{\text{knn}}_\ell(h) = \bar{c}^-_k(h) - \bar{c}^+_k(h).
\end{equation}
This is the only non-linear readout in our design. A safe-leaning 
prompt has smaller cosine distance to safe neighbors (small 
$\bar{c}^+_k$), yielding positive margin.

\paragraph{Supervised linear boundary.} For each layer $\ell$, we compute training statistics 
$\bar{h}_\ell, s_\ell \in \mathbb{R}^D$ and standardize 
$z = (h - \bar{h}_\ell) / s_\ell$. We fit an L2-regularized logistic 
regression on the standardized features, yielding weight $w_\ell$ and 
bias $b_\ell$; the linear margin is its signed score:
\begin{equation}
m^{\text{lin}}_\ell(h) = w_\ell^{\top} z + b_\ell.
\end{equation}
Positive margin indicates a safe-leaning representation under the 
per-layer supervised boundary.

\paragraph{Margin profiles across depth.} The three geometries provide complementary 
views of each layer's representation. Applied across all $L$ layers, 
they convert $H(x) \in \mathbb{R}^{L \times D}$ into three scalar 
margin trajectories
\begin{equation}
m^{G}(x) = \big[m^{G}_1(x),\, \ldots,\, m^{G}_L(x)\big] \in \mathbb{R}^L, 
\quad G \in \{\text{cent}, \text{knn}, \text{lin}\},
\end{equation}
reducing per-prompt dimensionality from $L \times D$ to $L \times 3$.

\subsection{Summarizing margin profiles with named axes}
\label{sec:method:features}

Geometry-Lite summarizes each margin profile along three named axes:
margin level, layer-to-layer change, and structural shape. These axes
separate three possible forms of layer-wise safety signal: persistent
boundary position, explicit across-layer motion, and coarse profile
shape. Throughout,
$L_{\mathrm{late}}=\{\ell:\ell>\lfloor 2L/3\rfloor\}$ denotes the final
third of layers; finite-difference summaries use the valid adjacent-layer
indices within this subset.

\paragraph{Margin level.}
The signed boundary position of the profile,
\[
\ell \longmapsto m^G_\ell(x),
\]
captures whether the prompt lies mostly on the safe or unsafe side of
the layer-wise boundary, and whether that position persists across
depth. This axis includes endpoint and extremal margins, negative-side
occupancy and area, net endpoint change, and the corresponding late-layer
boundary-position summaries.

\paragraph{Layer-to-layer change.}
The finite-difference signal,
\[
\ell \longmapsto \Delta m^G_\ell(x)
= m^G_{\ell+1}(x)-m^G_\ell(x),
\]
isolates local movement of the margin profile across adjacent layers,
separating across-layer motion from absolute boundary position. This
axis includes total movement toward the unsafe side across all layers
and the same quantity restricted to $L_{\mathrm{late}}$.

\paragraph{Structural shape.}
The coarse shape of the profile,
\[
m^G_{1:L}(x)
\quad\text{together with}\quad
\Delta m^G_{1:L-1}(x),
\]
captures whether the profile crosses the boundary, whether its path is
direct or oscillatory, and how much movement is required to reach its
endpoint. This axis includes first boundary crossing, variability of
layer-to-layer change, total path length, and a directional-consistency
ratio that is high for direct trajectories and low when movement cancels
across depth.

\medskip
\noindent
Applied to the centroid, $k$-NN, and linear margin profiles, these axes
produce a 39-dimensional summary
\[
\phi(x) = \mathrm{concat}\!\left(\phi^{\mathrm{cent}}(x),
\phi^{\mathrm{knn}}(x), \phi^{\mathrm{lin}}(x)\right) \in \mathbb{R}^{39},
\]
with $\phi^G(x) \in \mathbb{R}^{13}$ for each geometry. Precise feature
definitions and edge-case handling are in Appendix~\ref{app:feature_set}.
\subsection{Classification on interpretable summaries}
\label{sec:method:classifier}

The 39-dimensional feature vector $\phi(x)$ is fed to an 
L2-regularized logistic regression,
\begin{equation}
\hat{p}(y = 1 \mid x) = \sigma\big(w^{\top} \phi(x) + b\big), \quad w \in \mathbb{R}^{39},\ b \in \mathbb{R},
\end{equation}
trained by gradient descent on the training split (hyperparameters in 
Appendix~\ref{sec:method:setup}). The classifier has $39$ trainable weights plus a bias, with parameter count independent of the backbone and hidden dimension $D$.

\subsection{Multi-layer probe variants}
\label{sec:method:family}

We organize the multi-layer probing family by two axes: the per-layer
readout used to turn each hidden state into a scalar score, and the
cross-layer aggregation used to combine the resulting score profile.
The readout axis contrasts a label-informed class-mean direction with
an optimized supervised linear boundary. The aggregation axis contrasts
raw score stacking with named margin-profile summarization. Geometry-Lite
is the named-summarization member of this family, combining the profile
summaries of Section~\ref{sec:method:features} with the three geometric
readouts of Section~\ref{sec:method:geometries}.

\paragraph{MultiLayer-DIM.}
MultiLayer-DIM extends the difference-in-means (DIM) readout across
layers. Following \citet{marks2023geometry} and
\citet{zou2023representation}, each layer's hidden state is projected
onto the class-mean difference direction
\[
\hat{v}_\ell =
\frac{\mu^+_\ell-\mu^-_\ell}{\|\mu^+_\ell-\mu^-_\ell\|},
\]
and the resulting projection scores
\[
q_\ell = \hat{v}_\ell^\top h^{(\ell)}
\]
are aggregated by logistic regression across layers. This combines a
label-informed class-mean readout with raw cross-layer score stacking.

\paragraph{MultiLayer-Linear.} Each layer's hidden state is scored by
the supervised linear-boundary readout $m^{\text{lin}}_\ell$ from
Section~\ref{sec:method:geometries}, and the resulting scores are
aggregated by a top logistic regression,
\[
\hat{p}(y=1\mid x) =
\sigma\Big(\sum_{\ell=1}^L w_\ell\, m^{\text{lin}}_\ell + b\Big).
\]
This combines an optimized supervised readout with raw cross-layer score
stacking.

\medskip
\noindent
MultiLayer-DIM and MultiLayer-Linear share the same raw-stacking
aggregation and differ in the readout used at each layer. MultiLayer-Linear
is the closest raw-stacking counterpart to Geometry-Lite's supervised
linear-boundary branch, contrasting raw layer-wise score aggregation with
named margin-profile summaries.

\section{Empirical Experiments}
\label{sec:results}

We first evaluate prompt-level detection on the full in-distribution test
set (Section~\ref{sec:results:main}), then stress-test the same probes in
decision-critical regimes: low-false-positive operating points and a
predefined hard subset (Section~\ref{sec:results:critical}), and
leave-one-benchmark-out transfer (Section~\ref{sec:results:lobo}). These
experiments establish the performance context for the decomposition in
Section~\ref{sec:results:decomposition}.

\paragraph{Evaluation setup.}
We evaluate nine instruction-tuned backbones from the Llama, Gemma, and
Qwen3 families ($1.2$B--$70$B) on seven safety benchmarks: XSTest~\citep{rottger2024xstest}, WildJailbreak~\citep{jiang2024wildteaming},
JBB-Behaviors~\citep{chao2024jbb}, DoNotAnswer~\citep{wang2023donotanswer},
BeaverTails~\citep{ji2023beavertails}, PKU-SafeRLHF~\citep{ji2025pku},
and ToxicChat~\citep{lin2023toxicchat}. We use grouped stratified 70/30 splits and report
mean $\pm$ standard deviation over three seeds. We also define a hard
subset before method comparison, consisting of BeaverTails, ToxicChat,
and PKU-SafeRLHF, which emphasizes real-user, human-labeled, and
preference-style safety data. Baseline and hyperparameter details are in
Appendix~\ref{app:setup}.

\subsection{Main performance on the full test set}
\label{sec:results:main}

\begin{wrapfigure}{r}{0.48\columnwidth}
    \vspace{-12pt}
    \centering
    \includegraphics[width=0.46\columnwidth]{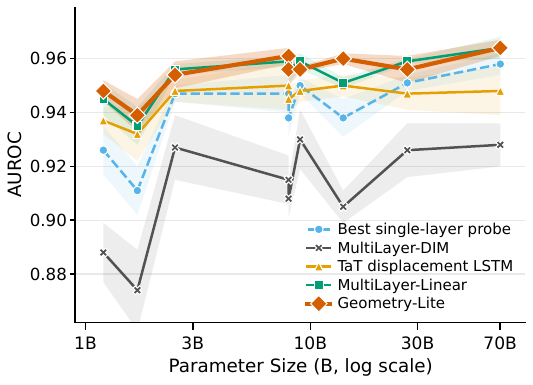}
    \caption{3-seed averaged Full-test AUROC across model backbones.}
    \label{fig:full_auroc_line}
    \vspace{-12pt}
\end{wrapfigure}

Figure~\ref{fig:full_auroc_line} shows full-test AUROC across model 
backbones. The two strongest pooled detectors are MultiLayer-Linear 
(0.954) and Geometry-Lite (0.955). They exceed the best single-layer 
probe by about 0.013--0.014, TaT-Disp-LSTM (0.945) by about 0.010, and 
the class-mean MultiLayer-DIM (0.911) by about 0.043. This ordering 
suggests that the in-distribution gain from multi-layer probing comes 
less from using many layers alone than from how each layer is converted 
into a safety-relevant score. MultiLayer-DIM has a pooled AUROC of 0.911, below the best single-layer probe at 0.941.

MultiLayer-Linear and Geometry-Lite have pooled AUROCs
of 0.954 and 0.955, respectively, and are the two highest-scoring
methods in Table~\ref{tab:full_main}. These two methods share a
supervised per-layer linear-boundary readout, while differing in how the
resulting layer-wise scores are summarized. Across individual backbones,
Geometry-Lite is highest or tied-highest on six backbones, while
MultiLayer-Linear is highest or tied-highest on five. Per-backbone
values are reported in Table~\ref{tab:full_main}.

\addvspace{1.3\baselineskip}
\begin{center}
\captionsetup{type=table}
\captionof{table}{AUROC on the full in-distribution test set across nine model backbones spanning three families (Llama-3, Gemma-2, Qwen3).}
\label{tab:full_main}
\normalsize
\setlength{\tabcolsep}{3pt}
\resizebox{\textwidth}{!}{%
\begin{tabular}{l|ccc|ccc|ccc|c}
\toprule
& \multicolumn{3}{c|}{\textbf{Llama}} & \multicolumn{3}{c|}{\textbf{Gemma}} & \multicolumn{3}{c|}{\textbf{Qwen3}} & \\
Method & 1B & 8B & 70B & 2B & 9B & 27B & 1.7B & 8B & 14B & Avg \\
\midrule
\multicolumn{11}{l}{\textit{Single-layer baselines}} \\
Final-layer probe       & .921$\pm$.008 & .947$\pm$.008 & .953$\pm$.005 & .933$\pm$.007 & .935$\pm$.001 & .936$\pm$.008 & .908$\pm$.005 & .933$\pm$.008 & .934$\pm$.003 & .933 \\
Best single-layer probe & .926$\pm$.009 & .947$\pm$.007 & .958$\pm$.004 & .947$\pm$.003 & .950$\pm$.003 & .951$\pm$.003 & .911$\pm$.009 & .938$\pm$.007 & .938$\pm$.007 & .941 \\
Refusal-style probe     & .901$\pm$.003 & .934$\pm$.009 & .949$\pm$.005 & .863$\pm$.014 & .898$\pm$.013 & .925$\pm$.012 & .898$\pm$.012 & .897$\pm$.016 & .900$\pm$.011 & .907 \\
\midrule
\multicolumn{11}{l}{\textit{Multi-layer methods}} \\
MultiLayer-DIM          & .888$\pm$.011 & .915$\pm$.009 & .928$\pm$.008 & .927$\pm$.012 & .930$\pm$.011 & .926$\pm$.010 & .874$\pm$.015 & .908$\pm$.007 & .905$\pm$.006 & .911 \\
TaT-Disp-LSTM           & .937$\pm$.005 & .950$\pm$.009 & .948$\pm$.009 & .948$\pm$.004 & .948$\pm$.004 & .947$\pm$.006 & .932$\pm$.010 & .945$\pm$.005 & .950$\pm$.004 & .945 \\
MultiLayer-Linear       & .945$\pm$.006 & .959$\pm$.002 & \textbf{.964$\pm$.004} & \textbf{.956$\pm$.002} & \textbf{.959$\pm$.003} & \textbf{.959$\pm$.001} & .935$\pm$.007 & \textbf{.956$\pm$.005} & .951$\pm$.002 & .954 \\
Geometry-Lite           & \textbf{.948$\pm$.004} & \textbf{.961$\pm$.003} & \textbf{.964$\pm$.003} & .954$\pm$.005 & .956$\pm$.001 & .956$\pm$.005 & \textbf{.939$\pm$.006} & \textbf{.956$\pm$.006} & \textbf{.960$\pm$.002} & .955 \\
\bottomrule
\end{tabular}%
}
\end{center}

\subsection{Utility in decision-critical regions}
\label{sec:results:critical}

Average AUROC on the full test set is dominated by easy, lexically 
obvious cases. We therefore evaluate low false-positive operating 
points and the benchmark-level hard subset defined above. Empirically, 
all methods drop by roughly $0.07$ AUROC on this subset, but the 
supervised multi-layer pair remains above the best single-layer probe 
(MultiLayer-Linear: $0.886 \pm 0.014$, Geometry-Lite: 
$0.883 \pm 0.015$, best single-layer: $0.868 \pm 0.021$; full table in 
Appendix~\ref{app:hard_full}).

\begin{wrapfigure}{r}{0.48\columnwidth}
    \vspace{-8pt}
    \centering
    \includegraphics[width=0.46\columnwidth]{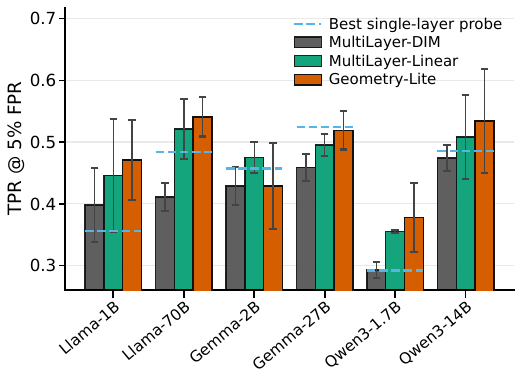}
    \caption{TPR@5\%FPR on the hard subset across six representative 
    backbones.}
    \label{fig:hard_tpr}
    \vspace{-8pt}
\end{wrapfigure}

Table~\ref{tab:tpr} reports TPR at 3\% and 5\% FPR on the full 
test set. At these operating points, the multi-layer advantage is 
larger than the AUROC gap suggests: Geometry-Lite improves over the 
best single-layer probe by $+0.079$ TPR at 3\% FPR and $+0.064$ at 
5\% FPR, despite a full-test AUROC gap of only $+0.014$. 
MultiLayer-Linear tracks Geometry-Lite closely at both thresholds.

On the hard subset, the same pattern holds qualitatively 
(Figure~\ref{fig:hard_tpr}). Geometry-Lite exceeds the best 
single-layer probe by $+0.049$ pooled TPR at 5\% FPR, while remaining 
close to MultiLayer-Linear. Appendix~\ref{app:tpr_full} reports 
per-backbone values, hard-subset values, and the 1\% FPR threshold; 
Appendix~\ref{app:static_uncertainty} reports a complementary static-probe uncertainty analysis, evaluating the bottom $q\%$ of prompts by $|p_{\mathrm{static}}(y{=}1\mid x)-0.5|$ for $q \in \{20,30,50\}$.

\vspace{-2pt}
\begin{center}
\captionsetup{type=table}
\captionof{table}{TPR at fixed false-positive rates on the full test 
set across nine model backbones. Pooled mean across nine 
backbones; per-backbone values, hard subset numbers, and 
1\% FPR threshold in Appendix~\ref{app:tpr_full}.}
\label{tab:tpr}
\footnotesize
\setlength{\tabcolsep}{8pt}
\begin{tabular}{lcc}
\toprule
Method & TPR@3\%FPR & TPR@5\%FPR \\
\midrule
Best single-layer probe & 0.564 & 0.686 \\
MultiLayer-DIM          & 0.466 & 0.574 \\
TaT-Disp-LSTM           & 0.567 & 0.694 \\
MultiLayer-Linear       & 0.641 & 0.747 \\
Geometry-Lite           & \textbf{0.643} & \textbf{0.750} \\
\bottomrule
\end{tabular}
\end{center}
\vspace{-8pt}

\subsection{Generalization under leave-one-benchmark-out (LOBO)}
\label{sec:results:lobo}

The full-test and hard-subset evaluations train and test on the same set of seven benchmarks. To assess generalization to unseen safety distributions, we adopt a leave-one-benchmark-out (LOBO) protocol. For each held-out benchmark, the probe is trained on the remaining six and evaluated only on the held-out benchmark. This complements concurrent dataset-level evaluations \citep{fomin2026benchmarks, damirchi2026truth}.

Table~\ref{tab:lobo} reports summarized per-held-out LOBO AUROC; the 
full per-backbone breakdown is provided in Appendix~\ref{app:lobo-breakdown}. 

\begin{center}
\captionsetup{type=table}
\captionof{table}{LOBO main results: AUROC per held-out benchmark, mean across nine model backbones. Best per row in bold; ties (within 0.005 AUROC) bolded jointly.}
\label{tab:lobo}
\footnotesize
\setlength{\tabcolsep}{4pt}
\resizebox{\textwidth}{!}{%
\begin{tabular}{l|c|cccc}
\toprule
& & \multicolumn{4}{c}{\textbf{Multi-layer methods}} \\
\cmidrule(l){3-6}
Held-out benchmark & Best single & MultiLayer-DIM & TaT-Disp-LSTM & MultiLayer-Linear & Geometry-Lite \\
\midrule
BeaverTails        & 0.754 & 0.733 & 0.742 & \textbf{0.767} & 0.763 \\
ToxicChat          & 0.751 & \textbf{0.877} & 0.818 & 0.799 & 0.798 \\
PKU-SafeRLHF       & 0.870 & \textbf{0.906} & 0.902 & 0.900 & 0.903 \\
JBB-Behaviors      & 0.832 & \textbf{0.937} & 0.879 & 0.871 & 0.862 \\
XSTest             & 0.975 & 0.979 & 0.988 & 0.990 & \textbf{0.996} \\
WildJailbreak      & 0.835 & 0.920 & 0.984 & 0.965 & \textbf{0.990} \\
DoNotAnswer        & 0.880 & 0.868 & 0.968 & 0.933 & \textbf{0.974} \\
\midrule
\textbf{Mean}      & 0.842 & 0.889 & 0.897 & 0.889 & \textbf{0.898} \\
\bottomrule
\end{tabular}%
}
\end{center}

In-distribution, MultiLayer-DIM sat 0.043 AUROC below the supervised
projection-based pair and below the best single-layer probe on every
backbone (Section~\ref{sec:results:main}). Under LOBO, however, this gap
largely disappears: MultiLayer-DIM (pooled 0.889) matches
MultiLayer-Linear (0.889) and exceeds the best single-layer probe
(0.842) on 46 of 63 (model, held-out) cells. MultiLayer-Linear and
Geometry-Lite remain close under LOBO as well. Geometry-Lite has the
highest pooled mean in Table~\ref{tab:lobo} (0.898), but the gap to
MultiLayer-Linear is small and is not the main LOBO finding.

The central change is the relative stability of class-mean readouts.
MultiLayer-DIM uses a label-informed class-mean direction rather than an
optimized per-layer boundary, suggesting that optimized boundaries are
sharper in-distribution while class-conditional mean geometry is more
stable under held-out benchmark shift. Section~\ref{sec:results:decomposition}
examines this trade-off directly.

\section{Decomposing the multi-layer safety signal}
\label{sec:results:decomposition}
We use Geometry-Lite to identify which form of multi-layer margin
evidence carries prompt-level safety signal. The construction separates
two choices that are entangled in raw multi-layer probes: how each layer
is scored, and how the resulting score profile is summarized. We
therefore ablate both axes across four regimes: the full test set, the
hard subset, all-LOBO transfer, and a hard-LOBO slice with held-out
benchmarks \{BeaverTails, ToxicChat, PKU-SafeRLHF\}.

\paragraph{Which summaries matter?}
The first ablation fixes all three readouts and varies the summaries
computed from each margin profile. We group the 13 summaries per geometry
into \emph{Magnitude} (7; endpoint, extremal, occupancy, and
boundary-relative area), \emph{Neg-drift} (2; finite-difference motion
toward the unsafe side), and \emph{Structural} (4; crossings and path
consistency).

\begin{table}[h]
\centering
\caption{Feature-group ablation across four evaluation regimes. ``Hard 
LOBO'' restricts held-out benchmarks to \{BeaverTails, ToxicChat, 
PKU-SafeRLHF\}; ``All LOBO'' averages over all seven held-out 
benchmarks. The final column reports TPR at 5\% FPR on the hard-LOBO 
slice. Best per column in bold (ties within $0.002$ for AUROC columns).}
\label{tab:feat_ablation}
\footnotesize
\setlength{\tabcolsep}{5pt}
\begin{tabular}{l|cccc|c}
\toprule
Feature subset & Full test & Hard subset & All LOBO & Hard LOBO 
& Hard LOBO TPR@5\%FPR \\
\midrule
Neg-drift only ($6$)               
& 0.900 & 0.823 & 0.795 & 0.781 & 0.272 \\
Structural only ($12$)             
& 0.851 & 0.767 & 0.776 & 0.690 & 0.167 \\
Magnitude only ($21$)              
& 0.954 & 0.881 & \textbf{0.898} & \textbf{0.821} & 0.298 \\
\midrule
Magnitude + Neg-drift ($27$)       
& \textbf{0.955} & \textbf{0.883} & \textbf{0.898} & \textbf{0.823} & \textbf{0.304} \\
Full (Geometry-Lite, $39$)         
& \textbf{0.955} & \textbf{0.883} & \textbf{0.898} & \textbf{0.821} & 0.291 \\
\bottomrule
\end{tabular}
\end{table}

Magnitude features alone match the full 39-feature Geometry-Lite probe 
within $0.001$--$0.002$ AUROC across all four regimes. Adding 
negative-drift features yields no measurable aggregate gain in any 
regime. Negative-drift alone is substantially weaker than magnitude 
alone (magnitude--drift gap of $0.040$--$0.103$ across regimes), and 
structural features alone are weaker still, trailing magnitude by 
$0.103$ on the full test, $0.114$ on the hard subset, $0.122$ on all 
LOBO, and $0.131$ on hard LOBO.
The strongest multi-layer signal is therefore not the local velocity 
of the representation from one layer to the next, but the persistence 
of the margin profile on one side of the safety boundary. Specifically, 
whether the prompt remains unsafe-like or safe-like across many layers, 
how far it is from the boundary at late layers, and how much of the 
profile occupies the negative half-plane.

\paragraph{Drift is sparse rather than aggregate.}
The feature ablation in Table~\ref{tab:feat_ablation} shows that
finite-difference summaries add little to aggregate AUROC, but two
further diagnostics qualify this picture. First, on the linear-boundary
margin, we compute standardized safe--unsafe separation for late-layer
margin levels and for their finite differences. On the hard-LOBO slice,
drift separation is much smaller than level separation across the 27
measured model--held-out cells, with a drift/level ratio of $11.7\%$ on
average, median $11.4\%$, and range $5.7$--$24.9\%$
(Appendix~\ref{app:snr_diagnostic}). Second, the low-FPR results in
Table~\ref{tab:feat_ablation} provide a narrower qualification: on hard
LOBO, negative-drift features slightly improve TPR@5\%FPR when added to
magnitude summaries (0.298 $\rightarrow$ 0.304). Sample-level inspection
also reveals occasional late-correction rescue cases, but these do not
translate into a large net gain when averaged across regimes
(Appendix~\ref{app:drift_cases_samples}). Thus, drift is best viewed as
a sparse correction axis that can aid recall under shifted low-FPR
thresholds, rather than as a primary source of aggregate separability.

\paragraph{Which geometric readouts matter?}
The second ablation holds the summaries fixed and varies which 
geometric readouts contribute to the margin profile.

\begin{table}[h]
\centering
\caption{Geometry ablation. Hard LOBO holds out
\{BeaverTails, ToxicChat, PKU-SafeRLHF\}; All LOBO averages over all
seven held-out benchmarks. The final column reports hard-LOBO
TPR@5\%FPR. Best AUROC values are bolded, with ties within $0.002$.}
\label{tab:geom_ablation}
\footnotesize
\setlength{\tabcolsep}{5pt}
\begin{tabular}{l|cccc|c}
\toprule
Geometry subset & Full test & Hard subset & All LOBO & Hard LOBO 
& Hard LOBO TPR@5\%FPR \\
\midrule
Centroid only        
& 0.883 & 0.836 & 0.874 & \textbf{0.837} & \textbf{0.375} \\
$k$-NN only          
& 0.942 & 0.858 & 0.874 & 0.789 & 0.202 \\
Linear-boundary only 
& 0.950 & 0.881 & 0.882 & 0.825 & 0.338 \\
\midrule
Centroid + $k$-NN    
& 0.941 & 0.852 & 0.864 & 0.773 & 0.158 \\
Centroid + linear    
& 0.951 & 0.882 & 0.890 & 0.829 & 0.346 \\
$k$-NN + linear      
& \textbf{0.955} & \textbf{0.885} & \textbf{0.900} & 0.824 & 0.312 \\
\midrule
All three (Geometry-Lite) 
& \textbf{0.955} & 0.883 & 0.898 & 0.821 & 0.291 \\
\bottomrule
\end{tabular}
\end{table}

Table~\ref{tab:geom_ablation} shows a regime-dependent pattern. 
In-distribution (full test and hard subset) and under mild benchmark 
shift (all-LOBO mean), the supervised linear-boundary readout is the 
strongest single branch, and the $k$-NN + linear-boundary pair 
matches the full three-readout model. Adding centroid distance on top 
of this pair yields no consistent gain. In these regimes, the 
class-conditional centroid readout is weaker than the optimized 
linear-boundary readout and behaves mostly as a redundant component of 
the full probe.

Under hard LOBO, however, the ordering changes. The centroid readout 
alone (0.837) becomes the strongest single branch, exceeding 
linear-boundary alone (0.825), $k$-NN alone (0.789), and the full 
three-readout probe (0.821). This stability also appears at the 
low-FPR operating point: centroid-only achieves the highest hard-LOBO 
TPR@5\%FPR (0.375), above linear-boundary only (0.338) and the full 
three-readout probe (0.291). Comparing all LOBO to hard LOBO, the 
$k$-NN + linear pair drops from 0.900 to 0.824 
($\Delta=-0.076$), whereas centroid alone drops only from 0.874 to 
0.837 ($\Delta=-0.037$). This pattern suggests a readout-level 
trade-off: optimized boundaries and local-neighborhood readouts are 
sharper in-distribution and under mild shift, while class-conditional 
mean geometry is more stable on the hardest held-out benchmarks. The 
same trade-off also appears in Section~\ref{sec:results:lobo}, where 
MultiLayer-DIM closes the gap to MultiLayer-Linear despite being weaker 
in-distribution.

\paragraph{Class-mean readouts preserve separation under shift.}
To examine this trade-off directly, we compute a held-out/train
separation ratio under LOBO. For each readout and layer, we compute a
standardized class-separation score, defined as the absolute
safe--unsafe mean margin gap normalized by within-class variability. We
average this score over late layers and report the held-out score divided
by the six-benchmark training score. This asks whether the class
separation defined by each readout remains visible after the held-out
benchmark changes the prompt distribution.

\begin{table}[h]
\centering
\caption{Readout-level held-out/train separation ratio under LOBO. We
compute late-layer standardized class separation before the final
aggregation classifier on the six LOBO training benchmarks and on the
held-out benchmark, then report their ratio. Values are means over
model--held-out cells; ratios are shown as mean $\pm$ standard error.
See Appendix~\ref{app:retention} for the exact definition.}
\label{tab:readout_retention}
\footnotesize
\setlength{\tabcolsep}{4pt}
\begin{tabular}{llcccc}
\toprule
Readout & Slice & Train sep. & Held-out sep. & Held-out/train & $n$ \\
\midrule
DIM      & All LOBO  & 1.035 & 1.168 & $1.192 \pm 0.108$ & 63 \\
Centroid & All LOBO  & 1.018 & 1.172 & $1.210 \pm 0.101$ & 63 \\
Linear boundary & All LOBO  & 2.703 & 1.139 & $0.427 \pm 0.023$ & 63 \\
\midrule
DIM      & Hard LOBO & 1.064 & 0.951 & $0.915 \pm 0.062$ & 27 \\
Centroid & Hard LOBO & 1.049 & 0.957 & $0.945 \pm 0.064$ & 27 \\
Linear boundary & Hard LOBO & 2.802 & 0.852 & $0.308 \pm 0.017$ & 27 \\
\bottomrule
\end{tabular}
\end{table}

Table~\ref{tab:readout_retention} shows that the optimized
linear-boundary readout is much sharper on the LOBO training side: its
train-side separation is roughly $2.6\times$ that of DIM or centroid.
However, this separation drops sharply under benchmark shift. Its
held-out/train ratio is 0.427 on all LOBO cells and 0.308 on the
hard-LOBO slice. In contrast, DIM and centroid have lower train-side
separation but preserve a larger fraction of it under shift, with ratios
around 1.2 on all LOBO cells and 0.9--0.95 on hard LOBO. This supports
the same interpretation as the geometry ablation: class-conditional mean
geometry is less sharp in-distribution but more stable under held-out
benchmark shift, whereas optimized linear boundaries capture stronger
training-mixture separation that transfers less reliably.

\section{Conclusion}
We introduced Geometry-Lite, a compact prompt-level probe that represents
multi-layer safety evidence through labeled margin-geometry axes. Across
nine instruction-tuned backbones and seven safety benchmarks, it improves
over single-layer probes while remaining competitive with raw multi-layer
score-stacking baselines.

Layer-wise safety evidence is concentrated in boundary-relative margin
position. Magnitude and boundary-occupancy summaries explain most
aggregate performance, while layer-to-layer motion is sparse and
contributes little to pooled AUROC. Drift can provide small
recall-oriented corrections under shifted low-FPR thresholds and in
localized late-correction cases, but is not a main source of aggregate
gains.

Benchmark shift changes which readout geometry is most stable. Optimized
linear boundaries are sharper in-distribution and under mild shift,
whereas class-mean readouts such as centroid geometry and DIM retain
separation more reliably on the hardest held-out benchmarks. Thus,
multi-layer safety probes should be evaluated by both in-distribution
AUROC and which geometric bias transfers under shift.

This study is limited to hand-designed margin summaries, single-turn
safety prompts, and binary safety labels. Future work could replace these
summaries with learned or theoretically motivated profile functionals,
extend the analysis to multi-turn safety monitoring, and apply this
margin-geometry view to other prompt-level tasks.

\bibliographystyle{unsrtnat}
\bibliography{references}


\appendix

\section{Feature set: precise definitions}
\label{app:feature_set}

This appendix defines the $13$ scalar summaries used for each margin
geometry $G\in\{\mathrm{cent},\mathrm{knn},\mathrm{lin}\}$. Concatenating
the three geometry-specific blocks gives the $39$-dimensional
Geometry-Lite representation.

\paragraph{Notation.}
For a fixed geometry $G$, write
\[
m(x)=\bigl(m_1(x),\ldots,m_L(x)\bigr)\in\mathbb{R}^L,
\qquad
\Delta m_\ell(x)=m_{\ell+1}(x)-m_\ell(x).
\]
Positive margins indicate the safe side of the boundary, and negative
margins indicate the unsafe side. A positive $\Delta m_\ell$ moves the
profile toward the safe side; a negative value moves it toward the
unsafe side.

We use the final third of layers as the late window,
\[
\mathcal{L}_{\mathrm{late}}
=
\{\ell \in \{1,\ldots,L\} : \ell > \lfloor 2L/3 \rfloor\}.
\]
For finite-difference features, the corresponding valid late drift
indices are
\[
\mathcal{D}_{\mathrm{late}}
=
\{\ell \in \{1,\ldots,L-1\} : \ell \in \mathcal{L}_{\mathrm{late}}
\ \text{and}\ \ell+1 \in \mathcal{L}_{\mathrm{late}}\}.
\]

\begin{table}[t]
\centering
\caption{The $13$ scalar summaries used for each geometry. MM denotes
features computed from $m_{1:L}$, and DR denotes features using
$\Delta m_{1:L-1}$.}
\label{tab:feature_set}
\small
\begin{tabular}{@{}lllc@{}}
\toprule
\textbf{Group} & \textbf{Feature} & \textbf{Definition} & \textbf{Late} \\
\midrule
\multirow{7}{*}{Margin level}
& \texttt{final\_margin}
& $m_L$
& \\

& \texttt{min\_margin}
& $\min_{\ell} m_\ell$
& \\

& \texttt{margin\_auc\_below\_zero}
& $\sum_{\ell=1}^{L}\max(0,-m_\ell)$
& \\

& \texttt{fraction\_below\_zero}
& $\frac{1}{L}\sum_{\ell=1}^{L}\mathbf{1}\{m_\ell<0\}$
& \\

& \texttt{drift\_mean}
& $\frac{1}{L-1}\sum_{\ell=1}^{L-1}\Delta m_\ell
   =\frac{m_L-m_1}{L-1}$
& \\

& \texttt{late\_min\_margin}
& $\min_{\ell\in\mathcal{L}_{\mathrm{late}}}m_\ell$
& \checkmark \\

& \texttt{late\_fraction\_below\_zero}
& $\frac{1}{|\mathcal{L}_{\mathrm{late}}|}
   \sum_{\ell\in\mathcal{L}_{\mathrm{late}}}
   \mathbf{1}\{m_\ell<0\}$
& \checkmark \\

\midrule
\multirow{2}{*}{Layer-to-layer}
& \texttt{cumulative\_negative\_drift}
& $\sum_{\ell=1}^{L-1}\max(0,-\Delta m_\ell)$
& \\

& \texttt{late\_negative\_drift}
& $\sum_{\ell\in\mathcal{D}_{\mathrm{late}}}
   \max(0,-\Delta m_\ell)$
& \checkmark \\

\midrule
\multirow{4}{*}{Structural}
& \texttt{first\_cross\_layer}
& $\min\{\ell:m_\ell<0\}$ if crossed, else $L$
& \\

& \texttt{drift\_std}
& $\mathrm{std}(\Delta m_1,\ldots,\Delta m_{L-1})$
& \\

& \texttt{margin\_progress\_total}
& $\sum_{\ell=1}^{L-1}|\Delta m_\ell|$
& \\

& \texttt{margin\_progress\_efficiency}
& $\frac{|m_L-m_1|}
        {\max(\varepsilon,\sum_{\ell=1}^{L-1}|\Delta m_\ell|)}$
& \\
\bottomrule
\end{tabular}
\end{table}

\paragraph{Numerical convention.}
We set $\varepsilon=10^{-8}$ in
\texttt{margin\_progress\_efficiency}. A constant profile therefore
receives value $0$ rather than producing an undefined ratio. All
features are deterministic functions of the fitted margin profile and
introduce no additional learned parameters.

\section{Experiments setup and Baselines}
\label{app:setup}

\subsection{Evaluation setup and baselines}
\label{sec:method:setup}

All activation-based methods access the final prompt-token hidden 
state across $L$ layers (Section~\ref{sec:method:problem}); the 
refusal-style probe is the only exception, following its original 
assistant-anchor protocol.

\paragraph{Models.} We evaluate on nine open-weight backbones spanning 
three families: Llama-3.2-1B, Llama-3.1-8B, Llama-3.1-70B 
\citep{grattafiori2024llama3}; Gemma-2-2B, Gemma-2-9B, Gemma-2-27B 
\citep{team2024gemma2}; and Qwen3-1.7B, Qwen3-8B, Qwen3-14B 
\citep{qwen2024qwen3}.

\paragraph{Splits and seeds.} We use grouped stratified 70/30 train/test 
splits, where paired or templated prompt variants (e.g., XSTest's 
safe/unsafe pairs) are kept within the same partition to prevent leakage. 
Stratification preserves benchmark and label proportions. All numbers 
report mean $\pm$ standard deviation across three random seeds.

\paragraph{Hyperparameters.} Per-layer linear-boundary classifier: 
L2-regularized logistic regression, $C=1.0$, liblinear solver, fit on 
standardized features. $k$-NN geometry: $k=8$, cosine distance. Final 
39-feature classifier: gradient-descent logistic regression with L2 
penalty coefficient $10^{-3}$, learning rate $0.1$, 600 epochs, fit on 
standardized features. Selection of the best-single-layer baseline uses 
inner validation on the training split.

\paragraph{Single-layer baselines.} 
\emph{Final-layer probe}: logistic regression on the last layer's hidden 
state. \emph{Best single-layer probe}: per-layer logistic regression, 
with the layer chosen by inner validation AUROC on the training split; 
for LOBO evaluation, layer selection uses only the six training 
benchmarks. \emph{Refusal-style probe} \citep{arditi2024refusal}: 
difference-in-means projection evaluated at the chat-template 
assistant-anchor position.

\paragraph{TaT-Disp-LSTM implementation.}
We use TaT-Disp-LSTM as a matched final-token adaptation of Truth as a
Trajectory (TaT)~\citep{damirchi2026truth}. To keep token-position access
matched to Geometry-Lite, MultiLayer-DIM, and MultiLayer-Linear, we fix
the input position to the final prompt token and let the LSTM process the
layer axis rather than the token axis. The input sequence is the
per-layer displacement
\[
\delta_\ell = h^{(\ell+1)} - h^{(\ell)}, \qquad \ell=1,\ldots,L-1,
\]
after train-side per-layer z-normalization. A single-layer unidirectional
LSTM with hidden size $256$ reads the length-$(L-1)$ displacement
sequence, and a linear head on the final LSTM state produces the logit.
We train with AdamW and BCEWithLogitsLoss using learning rate
$3\times10^{-4}$, batch size $32$, maximum $50$ epochs, and early
stopping with patience $5$ on a stratified $20\%$ validation split from
the training set. We use the same hyperparameters for all backbones,
seeds, and LOBO splits, without per-model tuning.

\subsection{Held-out/train separation ratio}
\label{app:retention}

This diagnostic is used only for post-hoc analysis of LOBO transfer. It
asks whether the class separation induced by each per-layer readout on
the six training benchmarks remains visible on the held-out benchmark.
We report a held-out/train separation ratio, defined as the standardized
class separation on the held-out benchmark divided by the corresponding
separation on the six-benchmark LOBO training set. Values near $1$
indicate comparable separation across train and held-out benchmarks,
values below $1$ indicate degradation under shift, and values above $1$
indicate stronger separation on the held-out benchmark under that
readout.

For each LOBO cell, readout $G$, and layer $\ell$, we compute a
standardized class-separation score
\begin{equation}
S^G_\ell(B)
=
\frac{
\left|
\mathbb{E}_{x\in B,\,y=0}\!\left[m^G_\ell(x)\right]
-
\mathbb{E}_{x\in B,\,y=1}\!\left[m^G_\ell(x)\right]
\right|
}{
\sqrt{
\operatorname{Var}_{x\in B,\,y=0}\!\left(m^G_\ell(x)\right)
+
\operatorname{Var}_{x\in B,\,y=1}\!\left(m^G_\ell(x)\right)
+
\epsilon
}
}.
\end{equation}
Here $B$ denotes either the six-benchmark LOBO training set or the
held-out benchmark, and we set $\epsilon=10^{-8}$. We average this
quantity over the late-layer window $L_{\mathrm{late}}$:
\begin{equation}
\bar{S}^{G}(B)
=
\frac{1}{|L_{\mathrm{late}}|}
\sum_{\ell\in L_{\mathrm{late}}}
S^G_\ell(B).
\end{equation}
The held-out/train separation ratio is
\begin{equation}
R^G
=
\frac{
\bar{S}^{G}(B_{\mathrm{heldout}})
}{
\bar{S}^{G}(B_{\mathrm{train}})
}.
\end{equation}
For DIM, $m^G_\ell$ denotes the difference-of-means projection score.
For centroid and linear boundary, $m^G_\ell$ denotes the corresponding
margin from Section~\ref{sec:method:geometries}.

\subsection{Hardware and Software Reproducibility}
\label{app:compute_classical}

Table~\ref{tab:software_versions} reports the hardware and software
stack used for all experiments.

\begin{table}[h]
\centering
\caption{Software and hardware stack used for all reported experiments.}
\label{tab:software_versions}
\tiny
\renewcommand{\arraystretch}{1.05}
\begin{tabular}{l|l}
\toprule
\textbf{Component} & \textbf{Version}  \\
\midrule
\multicolumn{2}{l}{\textit{Core scientific stack}} \\
Python                & 3.11.14       \\
NumPy                 & 2.4.1         \\
SciPy                 & 1.17.0        \\
pandas                & 2.3.3         \\
Matplotlib            & 3.10.8        \\
seaborn               & 0.13.2        \\
scikit-learn          & 1.8.0         \\

\midrule
\multicolumn{2}{l}{\textit{LLM stack (hidden-state extraction)}} \\
PyTorch               & 2.5.1 (CUDA 12.4)\\
HuggingFace Transformers & 5.6.0     \\
\texttt{accelerate}   & 1.13.0      \\
\texttt{bitsandbytes} & 0.49.2      \\
huggingface\_hub      & 1.11.0       \\
tokenizers            & 0.22.2       \\
safetensors           & 0.7.0        \\

\midrule
\multicolumn{2}{l}{\textit{Hardware}} \\
GPU                   & 2$\times$ Quadro RTX 8000 (48\,GB each)\\
NVIDIA driver / CUDA / cuDNN & 550.163.01 / 12.4 / 9.1.0 \\
OS                    & Linux 3.10 (CentOS 7, glibc 2.28) \\

\bottomrule
\end{tabular}
\end{table}

\section{Additional experimental results}
\subsection{Hard subset full results}
\label{app:hard_full}

Table~\ref{tab:hard_full} reports AUROC on the benchmark-level hard
subset, consisting of BeaverTails, ToxicChat, and PKU-SafeRLHF. This
subset emphasizes real-user, human-labeled, and preference-style safety
data, matching the subset defined in Section~\ref{sec:results}. The
supervised multi-layer pair remains strongest on average, with
MultiLayer-Linear and Geometry-Lite closely matched across backbones.

\begin{table}[h]
\centering
\caption{AUROC on the hard subset across nine model backbones. Values are
mean $\pm$ standard deviation over three seeds.}
\label{tab:hard_full}
\footnotesize
\setlength{\tabcolsep}{3pt}
\resizebox{\textwidth}{!}{%
\begin{tabular}{l|ccc|ccc|ccc}
\toprule
& \multicolumn{3}{c|}{\textbf{Llama}} 
& \multicolumn{3}{c|}{\textbf{Gemma}} 
& \multicolumn{3}{c}{\textbf{Qwen3}} \\
Method & 1B & 8B & 70B & 2B & 9B & 27B & 1.7B & 8B & 14B \\
\midrule
Final-layer probe       
& $.852 \pm .018$ & $.888 \pm .017$ & $.889 \pm .011$ 
& $.848 \pm .014$ & $.857 \pm .007$ & $.862 \pm .012$ 
& $.816 \pm .005$ & $.854 \pm .015$ & $.861 \pm .001$ \\

Best single-layer probe 
& $.848 \pm .014$ & $.875 \pm .018$ & $.890 \pm .005$ 
& $.877 \pm .007$ & $.878 \pm .014$ & $.888 \pm .001$ 
& $.825 \pm .015$ & $.860 \pm .006$ & $.874 \pm .010$ \\

Refusal-style probe     
& $.817 \pm .009$ & $.847 \pm .022$ & $.879 \pm .010$ 
& $.797 \pm .003$ & $.821 \pm .008$ & $.845 \pm .008$ 
& $.794 \pm .015$ & $.824 \pm .025$ & $.823 \pm .010$ \\

\midrule
MultiLayer-DIM          
& $.828 \pm .016$ & $.858 \pm .017$ & $.858 \pm .018$ 
& $.878 \pm .017$ & $.871 \pm .024$ & $.870 \pm .018$ 
& $.798 \pm .023$ & $.869 \pm .012$ & $.858 \pm .010$ \\

TaT-Disp-LSTM           
& $.862 \pm .007$ & $.887 \pm .014$ & $.876 \pm .018$ 
& $.874 \pm .004$ & $.876 \pm .006$ & $.874 \pm .009$ 
& $.839 \pm .011$ & $.873 \pm .010$ & $.878 \pm .004$ \\

MultiLayer-Linear       
& $.872 \pm .008$ & $.895 \pm .005$ & $\mathbf{.902 \pm .007}$ 
& $\mathbf{.887 \pm .005}$ & $\mathbf{.891 \pm .009}$ & $\mathbf{.895 \pm .002}$ 
& $\mathbf{.856 \pm .014}$ & $\mathbf{.888 \pm .012}$ & $.884 \pm .002$ \\

Geometry-Lite           
& $\mathbf{.875 \pm .001}$ & $\mathbf{.898 \pm .010}$ & $.898 \pm .008$ 
& $.878 \pm .012$ & $.886 \pm .004$ & $.888 \pm .011$ 
& $.851 \pm .011$ & $.882 \pm .012$ & $\mathbf{.894 \pm .006}$ \\
\bottomrule
\end{tabular}%
}
\end{table}
\subsection{TPR at low-FPR operating points}
\label{app:tpr_full}

Table~\ref{tab:tpr_full_per_backbone} expands the pooled full-test
threshold results from Table~\ref{tab:tpr}. 
Table~\ref{tab:tpr_hard_per_backbone} reports the corresponding
benchmark-defined hard-subset results used in Figure~\ref{fig:hard_tpr}.
Both tables report TPR at 1\%, 3\%, and 5\% FPR with mean $\pm$ std
over three seeds.

\begin{table}[h]
\centering
\scriptsize
\caption{\textbf{Full-test TPR at fixed false-positive rates.}
Per-backbone TPR at 1\%, 3\%, and 5\% FPR on the full test set. 
Values are mean $\pm$ std over three seeds for each backbone. 
The final column reports pooled mean $\pm$ std over all backbone--seed cells. 
Best values per backbone and FPR threshold are bolded; ties within 
$0.005$ are bolded jointly.}
\label{tab:tpr_full_per_backbone}
\resizebox{\textwidth}{!}{%
\begin{tabular}{llcccccccccc}
\toprule
FPR & Method & Llama-1B & Llama-8B & Llama-70B & Gemma-2B & Gemma-9B & Gemma-27B & Qwen3-1.7B & Qwen3-8B & Qwen3-14B & Avg \\
\midrule
\multirow{5}{*}{1\%} & Best single-layer & $.229_{\pm .048}$ & $.368_{\pm .041}$ & $.466_{\pm .114}$ & $\mathbf{.375_{\pm .084}}$ & $.329_{\pm .077}$ & $.369_{\pm .021}$ & $.206_{\pm .038}$ & $.338_{\pm .103}$ & $.390_{\pm .052}$ & $.341_{\pm .097}$ \\
 & MultiLayer-DIM & $.231_{\pm .077}$ & $.296_{\pm .058}$ & $.345_{\pm .053}$ & $.249_{\pm .033}$ & $.266_{\pm .053}$ & $.347_{\pm .047}$ & $.124_{\pm .036}$ & $.352_{\pm .015}$ & $.315_{\pm .019}$ & $.280_{\pm .081}$ \\
 & TaT-Disp-LSTM & $.308_{\pm .098}$ & $.371_{\pm .015}$ & $.337_{\pm .075}$ & $.356_{\pm .005}$ & $\mathbf{.396_{\pm .039}}$ & $.344_{\pm .045}$ & $.330_{\pm .013}$ & $.365_{\pm .062}$ & $.416_{\pm .089}$ & $.358_{\pm .058}$ \\
 & MultiLayer-Linear & $.326_{\pm .075}$ & $.438_{\pm .026}$ & $\mathbf{.486_{\pm .066}}$ & $.352_{\pm .077}$ & $.385_{\pm .072}$ & $\mathbf{.377_{\pm .015}}$ & $.297_{\pm .049}$ & $.411_{\pm .114}$ & $.388_{\pm .122}$ & $.385_{\pm .084}$ \\
 & Geometry-Lite & $\mathbf{.420_{\pm .038}}$ & $\mathbf{.481_{\pm .037}}$ & $.468_{\pm .087}$ & $.367_{\pm .084}$ & $.304_{\pm .072}$ & $.289_{\pm .084}$ & $\mathbf{.368_{\pm .048}}$ & $\mathbf{.474_{\pm .139}}$ & $\mathbf{.511_{\pm .052}}$ & $\mathbf{.409_{\pm .101}}$ \\
\midrule
\multirow{5}{*}{3\%} & Best single-layer & $.482_{\pm .055}$ & $.554_{\pm .051}$ & $.648_{\pm .096}$ & $.590_{\pm .080}$ & $.617_{\pm .064}$ & $.641_{\pm .066}$ & $.419_{\pm .064}$ & $.559_{\pm .039}$ & $.567_{\pm .009}$ & $.564_{\pm .089}$ \\
 & MultiLayer-DIM & $.406_{\pm .067}$ & $.475_{\pm .029}$ & $.484_{\pm .011}$ & $.500_{\pm .040}$ & $.495_{\pm .057}$ & $.470_{\pm .024}$ & $.380_{\pm .029}$ & $.472_{\pm .026}$ & $.511_{\pm .036}$ & $.466_{\pm .053}$ \\
 & TaT-Disp-LSTM & $.560_{\pm .041}$ & $.585_{\pm .038}$ & $.605_{\pm .023}$ & $.540_{\pm .008}$ & $.566_{\pm .031}$ & $.595_{\pm .044}$ & $.469_{\pm .034}$ & $.580_{\pm .079}$ & $.602_{\pm .020}$ & $.567_{\pm .052}$ \\
 & MultiLayer-Linear & $.571_{\pm .069}$ & $.642_{\pm .023}$ & $\mathbf{.716_{\pm .063}}$ & $\mathbf{.661_{\pm .018}}$ & $\mathbf{.688_{\pm .067}}$ & $\mathbf{.683_{\pm .035}}$ & $.524_{\pm .040}$ & $\mathbf{.638_{\pm .073}}$ & $.648_{\pm .034}$ & $\mathbf{.641_{\pm .071}}$ \\
 & Geometry-Lite & $\mathbf{.596_{\pm .037}}$ & $\mathbf{.675_{\pm .007}}$ & $.709_{\pm .016}$ & $.598_{\pm .063}$ & $.674_{\pm .062}$ & $\mathbf{.680_{\pm .008}}$ & $\mathbf{.542_{\pm .031}}$ & $\mathbf{.641_{\pm .046}}$ & $\mathbf{.671_{\pm .085}}$ & $\mathbf{.643_{\pm .065}}$ \\
\midrule
\multirow{5}{*}{5\%} & Best single-layer & $.633_{\pm .037}$ & $.704_{\pm .038}$ & $.748_{\pm .032}$ & $.707_{\pm .035}$ & $.719_{\pm .021}$ & $.753_{\pm .022}$ & $.554_{\pm .063}$ & $.656_{\pm .035}$ & $.704_{\pm .029}$ & $.687_{\pm .068}$ \\
 & MultiLayer-DIM & $.547_{\pm .042}$ & $.546_{\pm .004}$ & $.614_{\pm .029}$ & $.602_{\pm .043}$ & $.624_{\pm .031}$ & $.559_{\pm .029}$ & $.476_{\pm .020}$ & $.592_{\pm .023}$ & $.609_{\pm .007}$ & $.574_{\pm .051}$ \\
 & TaT-Disp-LSTM & $.677_{\pm .025}$ & $.713_{\pm .053}$ & $.716_{\pm .044}$ & $.683_{\pm .022}$ & $.681_{\pm .036}$ & $.706_{\pm .016}$ & $.628_{\pm .048}$ & $.716_{\pm .017}$ & $.723_{\pm .010}$ & $.694_{\pm .040}$ \\
 & MultiLayer-Linear & $\mathbf{.696_{\pm .038}}$ & $.787_{\pm .024}$ & $.774_{\pm .037}$ & $\mathbf{.757_{\pm .010}}$ & $\mathbf{.778_{\pm .025}}$ & $\mathbf{.794_{\pm .026}}$ & $.635_{\pm .061}$ & $\mathbf{.756_{\pm .035}}$ & $.745_{\pm .032}$ & $\mathbf{.747_{\pm .057}}$ \\
 & Geometry-Lite & $\mathbf{.692_{\pm .022}}$ & $\mathbf{.798_{\pm .048}}$ & $\mathbf{.791_{\pm .017}}$ & $.730_{\pm .037}$ & $\mathbf{.776_{\pm .011}}$ & $.777_{\pm .020}$ & $\mathbf{.657_{\pm .031}}$ & $\mathbf{.755_{\pm .005}}$ & $\mathbf{.776_{\pm .039}}$ & $\mathbf{.750_{\pm .052}}$ \\
\bottomrule
\end{tabular}}
\end{table}

\begin{table}[h]
\centering
\scriptsize
\caption{\textbf{Hard-subset TPR at fixed false-positive rates.}
Per-backbone TPR at 1\%, 3\%, and 5\% FPR on the benchmark-defined hard subset
(BeaverTails, ToxicChat, PKU-SafeRLHF). Values are mean $\pm$ std over
three seeds for each backbone. The final column reports pooled mean
$\pm$ std over all backbone--seed cells. Best values per backbone and
FPR threshold are bolded; ties within $0.005$ are bolded jointly.}
\label{tab:tpr_hard_per_backbone}
\resizebox{\textwidth}{!}{%
\begin{tabular}{llcccccccccc}
\toprule
FPR & Method & Llama-1B & Llama-8B & Llama-70B & Gemma-2B & Gemma-9B & Gemma-27B & Qwen3-1.7B & Qwen3-8B & Qwen3-14B & Avg \\
\midrule
\multirow{5}{*}{1\%} & Best single-layer & $.134_{\pm .020}$ & $.216_{\pm .073}$ & $.254_{\pm .044}$ & $\mathbf{.268_{\pm .048}}$ & $.158_{\pm .029}$ & $\mathbf{.258_{\pm .024}}$ & $.084_{\pm .060}$ & $.199_{\pm .127}$ & $.233_{\pm .041}$ & $.200_{\pm .079}$ \\
 & MultiLayer-DIM & $.124_{\pm .004}$ & $.213_{\pm .075}$ & $.239_{\pm .016}$ & $.200_{\pm .022}$ & $.149_{\pm .107}$ & $.241_{\pm .027}$ & $.047_{\pm .042}$ & $.272_{\pm .054}$ & $.207_{\pm .061}$ & $.188_{\pm .081}$ \\
 & TaT-Disp-LSTM & $.163_{\pm .096}$ & $\mathbf{.238_{\pm .087}}$ & $.234_{\pm .040}$ & $.202_{\pm .018}$ & $\mathbf{.179_{\pm .109}}$ & $.219_{\pm .066}$ & $.115_{\pm .027}$ & $.217_{\pm .109}$ & $\mathbf{.244_{\pm .099}}$ & $.201_{\pm .078}$ \\
 & MultiLayer-Linear & $.145_{\pm .046}$ & $.219_{\pm .052}$ & $.245_{\pm .041}$ & $.213_{\pm .070}$ & $.173_{\pm .029}$ & $.149_{\pm .052}$ & $.128_{\pm .005}$ & $.231_{\pm .090}$ & $.185_{\pm .102}$ & $.187_{\pm .065}$ \\
 & Geometry-Lite & $\mathbf{.268_{\pm .056}}$ & $.229_{\pm .018}$ & $\mathbf{.267_{\pm .010}}$ & $.180_{\pm .030}$ & $.145_{\pm .020}$ & $.122_{\pm .024}$ & $\mathbf{.146_{\pm .088}}$ & $\mathbf{.302_{\pm .157}}$ & $.200_{\pm .086}$ & $\mathbf{.206_{\pm .085}}$ \\
\midrule
\multirow{5}{*}{3\%} & Best single-layer & $.285_{\pm .138}$ & $.361_{\pm .038}$ & $.413_{\pm .057}$ & $.389_{\pm .084}$ & $.341_{\pm .033}$ & $.371_{\pm .079}$ & $.169_{\pm .036}$ & $.348_{\pm .104}$ & $.382_{\pm .084}$ & $.340_{\pm .097}$ \\
 & MultiLayer-DIM & $.354_{\pm .074}$ & $.360_{\pm .072}$ & $.347_{\pm .045}$ & $.310_{\pm .042}$ & $.287_{\pm .044}$ & $.407_{\pm .030}$ & $.124_{\pm .044}$ & $.398_{\pm .015}$ & $.388_{\pm .011}$ & $.330_{\pm .092}$ \\
 & TaT-Disp-LSTM & $.325_{\pm .146}$ & $.395_{\pm .056}$ & $.369_{\pm .044}$ & $.348_{\pm .029}$ & $\mathbf{.385_{\pm .026}}$ & $\mathbf{.438_{\pm .023}}$ & $.236_{\pm .014}$ & $.374_{\pm .086}$ & $.353_{\pm .070}$ & $.358_{\pm .078}$ \\
 & MultiLayer-Linear & $.328_{\pm .100}$ & $.374_{\pm .047}$ & $.424_{\pm .056}$ & $\mathbf{.419_{\pm .027}}$ & $.359_{\pm .022}$ & $.341_{\pm .045}$ & $.224_{\pm .046}$ & $.364_{\pm .162}$ & $.381_{\pm .092}$ & $.357_{\pm .087}$ \\
 & Geometry-Lite & $\mathbf{.371_{\pm .098}}$ & $\mathbf{.417_{\pm .036}}$ & $\mathbf{.441_{\pm .038}}$ & $.383_{\pm .108}$ & $.363_{\pm .060}$ & $.305_{\pm .056}$ & $\mathbf{.291_{\pm .055}}$ & $\mathbf{.408_{\pm .099}}$ & $\mathbf{.438_{\pm .056}}$ & $\mathbf{.380_{\pm .079}}$ \\
\midrule
\multirow{5}{*}{5\%} & Best single-layer & $.356_{\pm .085}$ & $.414_{\pm .023}$ & $.484_{\pm .041}$ & $.457_{\pm .068}$ & $.474_{\pm .063}$ & $\mathbf{.524_{\pm .066}}$ & $.292_{\pm .026}$ & $.426_{\pm .083}$ & $.485_{\pm .037}$ & $.435_{\pm .085}$ \\
 & MultiLayer-DIM & $.398_{\pm .060}$ & $.464_{\pm .048}$ & $.411_{\pm .022}$ & $.429_{\pm .031}$ & $.379_{\pm .010}$ & $.459_{\pm .022}$ & $.293_{\pm .013}$ & $.447_{\pm .005}$ & $.474_{\pm .021}$ & $.417_{\pm .060}$ \\
 & TaT-Disp-LSTM & $.407_{\pm .100}$ & $.500_{\pm .044}$ & $.502_{\pm .034}$ & $.402_{\pm .057}$ & $.468_{\pm .054}$ & $.473_{\pm .010}$ & $.341_{\pm .047}$ & $.455_{\pm .133}$ & $.487_{\pm .105}$ & $.448_{\pm .081}$ \\
 & MultiLayer-Linear & $.446_{\pm .092}$ & $.488_{\pm .027}$ & $.521_{\pm .049}$ & $\mathbf{.475_{\pm .025}}$ & $\mathbf{.535_{\pm .042}}$ & $.495_{\pm .018}$ & $.355_{\pm .002}$ & $\mathbf{.494_{\pm .101}}$ & $.508_{\pm .068}$ & $\mathbf{.480_{\pm .070}}$ \\
 & Geometry-Lite & $\mathbf{.471_{\pm .065}}$ & $\mathbf{.511_{\pm .039}}$ & $\mathbf{.541_{\pm .032}}$ & $.429_{\pm .070}$ & $.500_{\pm .039}$ & $.519_{\pm .031}$ & $\mathbf{.378_{\pm .056}}$ & $.471_{\pm .070}$ & $\mathbf{.534_{\pm .084}}$ & $\mathbf{.484_{\pm .070}}$ \\
\bottomrule
\end{tabular}}
\end{table}

\subsection{Full LOBO breakdown}
\label{app:lobo-breakdown}

This appendix supplements Section~\ref{sec:results:lobo} with the full
per-backbone LOBO breakdown. Each row corresponds to one 
(model, held-out benchmark) cell, and the benchmark-level mean rows match
the summarized LOBO table in the main text.

\begin{scriptsize}
\setlength{\tabcolsep}{2.5pt}
\renewcommand{\arraystretch}{0.92}

\begin{longtable}{@{}llccccc@{}}
\caption{Full LOBO AUROC breakdown across all methods. 
Abbreviations: BS = best single-layer probe, DIM = MultiLayer-DIM,
TaT = TaT-Disp-LSTM, ML = MultiLayer-Linear, GL = Geometry-Lite.}
\label{tab:lobo_full_breakdown}\\
\toprule
Bench. & Backbone & BS & DIM & TaT & ML & GL \\
\midrule
\endfirsthead
\multicolumn{7}{l}{\textit{BeaverTails}} \\
BeaverTails & Llama-1B & 0.727 & 0.694 & 0.715 & \textbf{0.744} & \textbf{0.747} \\
BeaverTails & Llama-8B & \textbf{0.773} & 0.719 & 0.728 & \textbf{0.776} & \textbf{0.772} \\
BeaverTails & Llama-70B & 0.777 & 0.730 & 0.739 & \textbf{0.794} & 0.781 \\
BeaverTails & Gemma-2B & 0.780 & 0.778 & \textbf{0.783} & 0.775 & \textbf{0.787} \\
BeaverTails & Gemma-9B & 0.766 & 0.764 & 0.769 & \textbf{0.781} & \textbf{0.777} \\
BeaverTails & Gemma-27B & 0.774 & 0.763 & 0.752 & \textbf{0.784} & 0.776 \\
BeaverTails & Qwen3-1.7B & 0.698 & 0.654 & 0.691 & \textbf{0.744} & 0.711 \\
BeaverTails & Qwen3-8B & \textbf{0.762} & 0.748 & \textbf{0.762} & \textbf{0.759} & \textbf{0.761} \\
BeaverTails & Qwen3-14B & 0.732 & 0.743 & 0.742 & 0.749 & \textbf{0.755} \\
\textit{Mean} & -- & 0.754 & 0.733 & 0.742 & \textbf{0.767} & \textbf{0.763} \\
\midrule
\multicolumn{7}{l}{\textit{ToxicChat}} \\
ToxicChat & Llama-1B & 0.798 & \textbf{0.867} & 0.825 & 0.796 & 0.805 \\
ToxicChat & Llama-8B & 0.752 & \textbf{0.880} & 0.858 & 0.816 & 0.807 \\
ToxicChat & Llama-70B & 0.847 & \textbf{0.887} & \textbf{0.887} & 0.827 & 0.857 \\
ToxicChat & Gemma-2B & 0.733 & \textbf{0.900} & 0.792 & 0.795 & 0.783 \\
ToxicChat & Gemma-9B & 0.711 & \textbf{0.883} & 0.796 & 0.786 & 0.782 \\
ToxicChat & Gemma-27B & 0.760 & \textbf{0.853} & 0.816 & 0.804 & 0.811 \\
ToxicChat & Qwen3-1.7B & 0.716 & \textbf{0.823} & 0.773 & 0.780 & 0.773 \\
ToxicChat & Qwen3-8B & 0.700 & \textbf{0.913} & 0.783 & 0.778 & 0.742 \\
ToxicChat & Qwen3-14B & 0.739 & \textbf{0.891} & 0.829 & 0.804 & 0.820 \\
\textit{Mean} & -- & 0.751 & \textbf{0.877} & 0.818 & 0.799 & 0.798 \\
\midrule
\multicolumn{7}{l}{\textit{PKU-SafeRLHF}} \\
PKU-SafeRLHF & Llama-1B & 0.833 & \textbf{0.893} & \textbf{0.898} & 0.890 & 0.890 \\
PKU-SafeRLHF & Llama-8B & 0.886 & 0.906 & 0.907 & \textbf{0.919} & \textbf{0.917} \\
PKU-SafeRLHF & Llama-70B & 0.891 & 0.902 & \textbf{0.935} & 0.926 & 0.919 \\
PKU-SafeRLHF & Gemma-2B & 0.875 & \textbf{0.902} & 0.887 & \textbf{0.902} & \textbf{0.905} \\
PKU-SafeRLHF & Gemma-9B & 0.884 & \textbf{0.916} & 0.908 & \textbf{0.914} & 0.909 \\
PKU-SafeRLHF & Gemma-27B & 0.891 & \textbf{0.932} & 0.926 & 0.923 & \textbf{0.931} \\
PKU-SafeRLHF & Qwen3-1.7B & 0.832 & \textbf{0.861} & 0.828 & \textbf{0.860} & \textbf{0.857} \\
PKU-SafeRLHF & Qwen3-8B & 0.864 & \textbf{0.925} & 0.915 & 0.883 & 0.890 \\
PKU-SafeRLHF & Qwen3-14B & 0.871 & \textbf{0.917} & \textbf{0.914} & 0.879 & 0.907 \\
\textit{Mean} & -- & 0.870 & \textbf{0.906} & \textbf{0.902} & 0.900 & \textbf{0.903} \\
\midrule
\multicolumn{7}{l}{\textit{JBB-Behaviors}} \\
JBB-Behaviors & Llama-1B & 0.756 & \textbf{0.914} & 0.844 & 0.822 & 0.795 \\
JBB-Behaviors & Llama-8B & 0.846 & \textbf{0.932} & 0.893 & 0.884 & 0.867 \\
JBB-Behaviors & Llama-70B & 0.922 & \textbf{0.949} & 0.894 & 0.913 & 0.900 \\
JBB-Behaviors & Gemma-2B & 0.816 & \textbf{0.946} & 0.890 & 0.888 & 0.869 \\
JBB-Behaviors & Gemma-9B & 0.859 & \textbf{0.943} & 0.894 & 0.882 & 0.872 \\
JBB-Behaviors & Gemma-27B & 0.847 & \textbf{0.965} & 0.858 & 0.888 & 0.874 \\
JBB-Behaviors & Qwen3-1.7B & 0.753 & \textbf{0.921} & 0.846 & 0.819 & 0.801 \\
JBB-Behaviors & Qwen3-8B & 0.870 & \textbf{0.934} & 0.910 & 0.881 & 0.882 \\
JBB-Behaviors & Qwen3-14B & 0.821 & \textbf{0.929} & 0.888 & 0.858 & 0.897 \\
\textit{Mean} & -- & 0.832 & \textbf{0.937} & 0.879 & 0.871 & 0.862 \\
\midrule
\multicolumn{7}{l}{\textit{XSTest}} \\
XSTest & Llama-1B & 0.962 & 0.962 & 0.973 & 0.985 & \textbf{0.992} \\
XSTest & Llama-8B & 0.975 & 0.992 & 0.991 & 0.993 & \textbf{0.999} \\
XSTest & Llama-70B & 0.990 & \textbf{0.997} & \textbf{0.998} & \textbf{0.998} & \textbf{0.999} \\
XSTest & Gemma-2B & 0.965 & 0.986 & \textbf{0.992} & 0.990 & \textbf{0.995} \\
XSTest & Gemma-9B & 0.967 & \textbf{0.996} & 0.990 & \textbf{0.995} & \textbf{0.998} \\
XSTest & Gemma-27B & 0.988 & 0.992 & 0.994 & \textbf{0.997} & \textbf{1.000} \\
XSTest & Qwen3-1.7B & 0.960 & 0.938 & 0.963 & 0.967 & \textbf{0.985} \\
XSTest & Qwen3-8B & 0.989 & 0.985 & \textbf{0.995} & 0.992 & \textbf{0.998} \\
XSTest & Qwen3-14B & 0.975 & 0.963 & \textbf{0.996} & 0.991 & \textbf{0.998} \\
\textit{Mean} & -- & 0.974 & 0.979 & 0.988 & 0.990 & \textbf{0.996} \\
\midrule
\multicolumn{7}{l}{\textit{WildJailbreak}} \\
WildJailbreak & Llama-1B & 0.873 & 0.890 & 0.959 & 0.964 & \textbf{0.995} \\
WildJailbreak & Llama-8B & 0.889 & 0.928 & \textbf{0.995} & 0.991 & \textbf{0.998} \\
WildJailbreak & Llama-70B & 0.949 & 0.962 & 0.985 & 0.973 & \textbf{0.995} \\
WildJailbreak & Gemma-2B & 0.823 & 0.914 & 0.975 & 0.965 & \textbf{0.992} \\
WildJailbreak & Gemma-9B & 0.859 & 0.930 & \textbf{0.982} & \textbf{0.981} & \textbf{0.984} \\
WildJailbreak & Gemma-27B & 0.827 & 0.904 & \textbf{0.988} & 0.981 & 0.979 \\
WildJailbreak & Qwen3-1.7B & 0.600 & 0.925 & \textbf{0.993} & 0.905 & 0.979 \\
WildJailbreak & Qwen3-8B & 0.840 & 0.898 & 0.987 & 0.989 & \textbf{0.995} \\
WildJailbreak & Qwen3-14B & 0.866 & 0.932 & \textbf{0.997} & 0.940 & \textbf{0.995} \\
\textit{Mean} & -- & 0.836 & 0.920 & 0.984 & 0.965 & \textbf{0.990} \\
\midrule
\multicolumn{7}{l}{\textit{DoNotAnswer}} \\
DoNotAnswer & Llama-1B & 0.789 & 0.785 & \textbf{0.949} & 0.903 & \textbf{0.946} \\
DoNotAnswer & Llama-8B & 0.935 & 0.877 & 0.965 & 0.964 & \textbf{0.983} \\
DoNotAnswer & Llama-70B & 0.885 & 0.924 & \textbf{0.980} & 0.965 & \textbf{0.980} \\
DoNotAnswer & Gemma-2B & 0.909 & 0.911 & 0.975 & 0.962 & \textbf{0.989} \\
DoNotAnswer & Gemma-9B & 0.874 & 0.907 & \textbf{0.976} & 0.944 & \textbf{0.975} \\
DoNotAnswer & Gemma-27B & 0.907 & 0.903 & \textbf{0.978} & 0.932 & \textbf{0.979} \\
DoNotAnswer & Qwen3-1.7B & 0.816 & 0.832 & 0.948 & 0.874 & \textbf{0.956} \\
DoNotAnswer & Qwen3-8B & 0.908 & 0.815 & 0.966 & 0.928 & \textbf{0.978} \\
DoNotAnswer & Qwen3-14B & 0.904 & 0.856 & \textbf{0.974} & 0.926 & \textbf{0.976} \\
\textit{Mean} & -- & 0.881 & 0.868 & 0.968 & 0.933 & \textbf{0.974} \\
\midrule
\textbf{Overall mean} & -- & 0.843 & 0.889 & \textbf{0.897} & 0.889 & \textbf{0.898} \\
\bottomrule
\end{longtable}
\end{scriptsize}

\subsection{Low FPR performance in Hard LOBO}

\begin{table}[h]
\centering
\caption{Hard-LOBO TPR at fixed false-positive rates for feature-group
ablations. Values are pooled means over 27 model--held-out cells.}
\label{tab:hard_lobo_lowfpr_feature}
\footnotesize
\begin{tabular}{lccc}
\toprule
Feature subset & TPR@1\%FPR & TPR@3\%FPR & TPR@5\%FPR \\
\midrule
Structural only          & 0.033 & 0.103 & 0.167 \\
Negative-drift only      & 0.071 & 0.197 & 0.272 \\
Magnitude only           & \textbf{0.103} & 0.209 & 0.298 \\
Magnitude + Neg-drift    & \textbf{0.103} & \textbf{0.216} & \textbf{0.304} \\
Full Geometry-Lite       & 0.097 & 0.203 & 0.291 \\
\bottomrule
\end{tabular}
\end{table}

\section{Additional Experiments}
\subsection{Static-probe uncertainty slice}
\label{app:static_uncertainty}

To complement the benchmark-defined hard subset in
Section~\ref{sec:results:critical}, we evaluate a static-probe
uncertainty slice. For each backbone and seed, we rank test prompts by
$|p_{\mathrm{static}}(y{=}1\mid x)-0.5|$, where $p_{\mathrm{static}}$
is the validation-selected best single-layer probe, and select the
bottom $q\%$. Unless otherwise stated, we use $q=30$. This slice is not
used for model selection; it asks whether the ordering observed on hard
benchmarks also holds on prompts where the static probe is least
decisive.

Table~\ref{tab:static_uncertainty_auroc} reports AUROC on this slice.
The supervised multi-layer pair remains strongest, with
MultiLayer-Linear and Geometry-Lite nearly tied on average
($0.827$ vs. $0.826$). Both are substantially above the best
single-layer probe ($0.758$) and TaT-Disp-LSTM ($0.794$). Thus, this
diagnostic supports the hard-subset result without introducing a new
ranking: the gain in difficult regions comes from supervised
multi-layer boundary evidence, while Geometry-Lite remains matched to
its raw-stacking counterpart.

\begin{table}[h]
\centering
\scriptsize
\caption{\textbf{Static-probe uncertainty slice, AUROC.}
AUROC on the bottom $30\%$ of test prompts ranked by
$|p_{\mathrm{static}}(y{=}1\mid x)-0.5|$, computed separately for each
backbone and seed. Best values per column are bolded; ties within
$0.005$ are bolded jointly.}
\label{tab:static_uncertainty_auroc}
\resizebox{\textwidth}{!}{
\begin{tabular}{lcccccccccc}
\toprule
Method & Llama-1B & Llama-8B & Llama-70B & Gemma-2B & Gemma-9B & Gemma-27B & Qwen3-1.7B & Qwen3-8B & Qwen3-14B & Avg \\
\midrule
Final-layer probe & .761 & .789 & .800 & .749 & .750 & .755 & .750 & .728 & .771 & .761 \\
Best single-layer probe & .735 & .769 & .806 & .757 & .773 & .763 & .722 & .749 & .748 & .758 \\
Refusal-style probe & .745 & .763 & .799 & .696 & .729 & .731 & .757 & .735 & .709 & .741 \\
MultiLayer-DIM & .699 & .703 & .720 & .766 & .752 & .725 & .698 & .697 & .723 & .720 \\
TaT-Disp-LSTM & .798 & .783 & .774 & .791 & .781 & .774 & .823 & .795 & .825 & .794 \\
MultiLayer-Linear & \textbf{.830} & .829 & \textbf{.836} & \textbf{.820} & \textbf{.823} & \textbf{.819} & \textbf{.838} & \textbf{.824} & .827 & \textbf{.827} \\
Geometry-Lite & \textbf{.826} & \textbf{.834} & \textbf{.837} & .810 & .810 & .807 & \textbf{.838} & \textbf{.822} & \textbf{.851} & \textbf{.826} \\
\bottomrule
\end{tabular}}
\end{table}

Table~\ref{tab:static_uncertainty_tpr} reports the same diagnostic at a
low false-positive operating point. At TPR at 1\%, 3\%, 5\% FPR, Geometry-Lite and
MultiLayer-Linear again form the strongest pair.
\begin{table}[h]
\centering
\scriptsize
\caption{\textbf{Static-probe uncertainty slice, TPR at fixed FPRs.}
TPR at 1\%, 3\%, and 5\% false-positive rate on the bottom $30\%$ of
test prompts ranked by $|p_{\mathrm{static}}(y{=}1\mid x)-0.5|$.
Values are mean $\pm$ std over three seeds for each backbone. The final
column reports pooled mean $\pm$ std over all backbone--seed cells.
Best values per column and FPR threshold are bolded; ties within
$0.005$ are bolded jointly.}
\label{tab:static_uncertainty_tpr}
\resizebox{\textwidth}{!}{
\begin{tabular}{llcccccccccc}
\toprule
FPR & Method & Llama-1B & Llama-8B & Llama-70B & Gemma-2B & Gemma-9B & Gemma-27B & Qwen3-1.7B & Qwen3-8B & Qwen3-14B & Avg \\
\midrule
\multirow{7}{*}{1\%}
& Final-layer probe
& $.110{\pm}.089$ & $.094{\pm}.064$ & $.085{\pm}.129$
& $.035{\pm}.018$ & $.022{\pm}.007$ & $.038{\pm}.048$
& $.028{\pm}.012$ & $.099{\pm}.073$ & $.060{\pm}.042$
& $.063{\pm}.064$ \\
& Best single-layer probe
& $.059{\pm}.042$ & $.076{\pm}.054$ & $.087{\pm}.084$
& $.094{\pm}.032$ & $.071{\pm}.039$ & $.092{\pm}.079$
& $.041{\pm}.036$ & $.091{\pm}.034$ & $.080{\pm}.087$
& $.077{\pm}.051$ \\
& Refusal-style probe
& $.065{\pm}.015$ & $.059{\pm}.011$ & $.131{\pm}.066$
& $.055{\pm}.026$ & $.007{\pm}.007$ & $.081{\pm}.038$
& $.050{\pm}.005$ & $.024{\pm}.008$ & $.065{\pm}.032$
& $.060{\pm}.042$ \\
& MultiLayer-DIM
& $.135{\pm}.106$ & $.072{\pm}.032$ & $.134{\pm}.060$
& $.053{\pm}.033$ & $.046{\pm}.054$ & $.100{\pm}.085$
& $.098{\pm}.112$ & $.057{\pm}.046$ & $.088{\pm}.046$
& $.087{\pm}.066$ \\
& TaT-Disp-LSTM
& $.153{\pm}.047$ & $.096{\pm}.066$ & $.082{\pm}.031$
& $.076{\pm}.060$ & $\mathbf{.168{\pm}.027}$ & $\mathbf{.105{\pm}.092}$
& $\mathbf{.205{\pm}.057}$ & $.114{\pm}.019$ & $.102{\pm}.093$
& $.122{\pm}.065$ \\
& MultiLayer-Linear
& $.146{\pm}.082$ & $.126{\pm}.069$ & $.112{\pm}.079$
& $\mathbf{.110{\pm}.080}$ & $.112{\pm}.077$ & $\mathbf{.107{\pm}.142}$
& $.128{\pm}.100$ & $.104{\pm}.077$ & $.167{\pm}.141$
& $.124{\pm}.084$ \\
& Geometry-Lite
& $\mathbf{.173{\pm}.025}$ & $\mathbf{.198{\pm}.021}$ & $\mathbf{.154{\pm}.156}$
& $.088{\pm}.056$ & $\mathbf{.167{\pm}.008}$ & $\mathbf{.109{\pm}.151}$
& $.169{\pm}.036$ & $\mathbf{.168{\pm}.102}$ & $\mathbf{.209{\pm}.133}$
& $\mathbf{.159{\pm}.087}$ \\
\midrule
\multirow{7}{*}{3\%}
& Final-layer probe
& $.150{\pm}.117$ & $.134{\pm}.070$ & $.195{\pm}.169$
& $.080{\pm}.019$ & $.076{\pm}.006$ & $.087{\pm}.068$
& $.148{\pm}.034$ & $.186{\pm}.127$ & $.116{\pm}.033$
& $.130{\pm}.085$ \\
& Best single-layer probe
& $.171{\pm}.026$ & $.155{\pm}.041$ & $.203{\pm}.067$
& $.172{\pm}.071$ & $.172{\pm}.041$ & $.193{\pm}.070$
& $.086{\pm}.073$ & $.141{\pm}.067$ & $.126{\pm}.063$
& $.158{\pm}.061$ \\
& Refusal-style probe
& $.150{\pm}.055$ & $.109{\pm}.049$ & $.253{\pm}.164$
& $.105{\pm}.011$ & $.067{\pm}.043$ & $.130{\pm}.045$
& $.136{\pm}.067$ & $.074{\pm}.036$ & $.142{\pm}.043$
& $.130{\pm}.078$ \\
& MultiLayer-DIM
& $.170{\pm}.090$ & $.130{\pm}.021$ & $.179{\pm}.055$
& $.133{\pm}.037$ & $.142{\pm}.041$ & $.159{\pm}.033$
& $.156{\pm}.115$ & $.124{\pm}.019$ & $.160{\pm}.052$
& $.150{\pm}.053$ \\
& TaT-Disp-LSTM
& $.208{\pm}.006$ & $.155{\pm}.048$ & $.189{\pm}.098$
& $.113{\pm}.080$ & $.224{\pm}.026$ & $\mathbf{.239{\pm}.057}$
& $\mathbf{.294{\pm}.117}$ & $.136{\pm}.032$ & $.160{\pm}.077$
& $.191{\pm}.079$ \\
& MultiLayer-Linear
& $.202{\pm}.115$ & $.252{\pm}.060$ & $\mathbf{.279{\pm}.118}$
& $\mathbf{.213{\pm}.069}$ & $\mathbf{.241{\pm}.149}$ & $.227{\pm}.109$
& $.219{\pm}.058$ & $.216{\pm}.097$ & $.245{\pm}.099$
& $.233{\pm}.087$ \\
& Geometry-Lite
& $\mathbf{.248{\pm}.038}$ & $\mathbf{.291{\pm}.026}$ & $.258{\pm}.153$
& $.153{\pm}.091$ & $\mathbf{.241{\pm}.023}$ & $.221{\pm}.133$
& $.247{\pm}.039$ & $\mathbf{.242{\pm}.033}$ & $\mathbf{.271{\pm}.157}$
& $\mathbf{.241{\pm}.086}$ \\
\midrule
\multirow{7}{*}{5\%}
& Final-layer probe
& $.183{\pm}.129$ & $.198{\pm}.117$ & $.289{\pm}.217$
& $.119{\pm}.024$ & $.159{\pm}.034$ & $.128{\pm}.099$
& $.211{\pm}.022$ & $.303{\pm}.035$ & $.219{\pm}.078$
& $.201{\pm}.106$ \\
& Best single-layer probe
& $.263{\pm}.037$ & $.210{\pm}.077$ & $.263{\pm}.045$
& $.228{\pm}.054$ & $.248{\pm}.085$ & $.283{\pm}.085$
& $.133{\pm}.077$ & $.215{\pm}.057$ & $.188{\pm}.060$
& $.226{\pm}.071$ \\
& Refusal-style probe
& $.211{\pm}.031$ & $.190{\pm}.048$ & $.340{\pm}.198$
& $.177{\pm}.014$ & $.148{\pm}.051$ & $.195{\pm}.036$
& $.189{\pm}.056$ & $.162{\pm}.026$ & $.198{\pm}.023$
& $.201{\pm}.082$ \\
& MultiLayer-DIM
& $.213{\pm}.069$ & $.200{\pm}.008$ & $.212{\pm}.042$
& $.265{\pm}.073$ & $.246{\pm}.124$ & $.205{\pm}.031$
& $.194{\pm}.084$ & $.191{\pm}.014$ & $.224{\pm}.069$
& $.217{\pm}.060$ \\
& TaT-Disp-LSTM
& $.276{\pm}.019$ & $.213{\pm}.073$ & $.269{\pm}.074$
& $.240{\pm}.032$ & $.330{\pm}.027$ & $.300{\pm}.054$
& $\mathbf{.346{\pm}.091}$ & $.290{\pm}.078$ & $.271{\pm}.145$
& $.282{\pm}.074$ \\
& MultiLayer-Linear
& $\mathbf{.331{\pm}.100}$ & $.317{\pm}.094$ & $.340{\pm}.074$
& $\mathbf{.320{\pm}.033}$ & $.330{\pm}.080$ & $\mathbf{.349{\pm}.053}$
& $.314{\pm}.170$ & $\mathbf{.373{\pm}.074}$ & $.292{\pm}.101$
& $.330{\pm}.081$ \\
& Geometry-Lite
& $\mathbf{.326{\pm}.056}$ & $\mathbf{.378{\pm}.095}$ & $\mathbf{.362{\pm}.129}$
& $.264{\pm}.018$ & $\mathbf{.344{\pm}.028}$ & $.340{\pm}.064$
& $.315{\pm}.056$ & $.319{\pm}.036$ & $\mathbf{.370{\pm}.145}$
& $\mathbf{.335{\pm}.075}$ \\
\bottomrule
\end{tabular}}
\end{table}

\begin{figure}[h!]
    \centering
    \includegraphics[width=\textwidth]{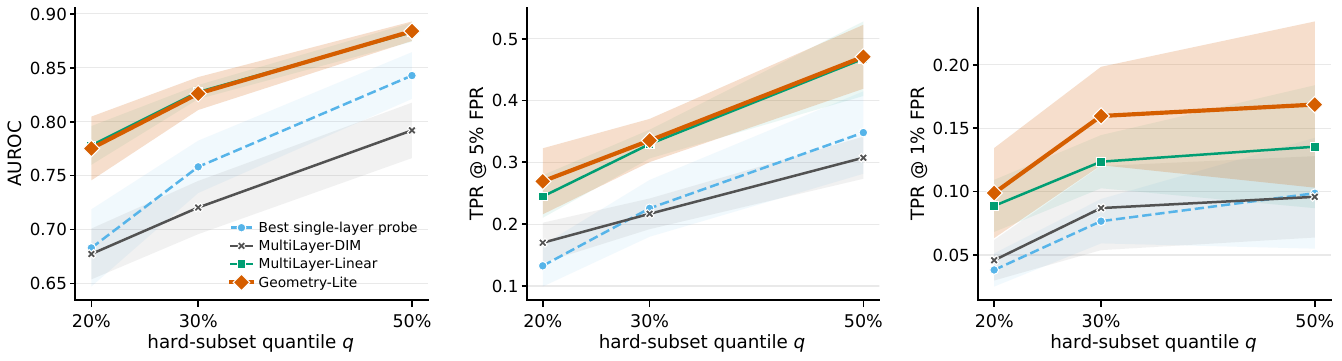}
    \caption{\textbf{Sensitivity to uncertainty-slice size.}
    We vary $q$, the fraction of test prompts closest to the
    validation-selected best single-layer probe's probability boundary,
    as measured by $|p_{\mathrm{static}}(y{=}1\mid x)-0.5|$. The
    ordering remains stable across slice sizes.}
    \label{fig:static_uncertainty_q}
\end{figure}

Finally, Figure~\ref{fig:static_uncertainty_q} varies the slice size
$q$. The qualitative ordering is stable: supervised multi-layer methods
remain above single-layer probes, while Geometry-Lite remains closely
matched to MultiLayer-Linear. This confirms that the diagnostic is not
specific to the default $q=30$ threshold.

\newpage
\subsection{Margin-level versus drift class separation}
\label{app:snr_diagnostic}

Section~\ref{sec:results:decomposition} uses a post-hoc class-separation
diagnostic to compare the linear-boundary margin profile with its finite
differences. This diagnostic is not used for training. It is computed
only to test whether layer-to-layer change features carry class
information comparable to margin-level features.

We use only the supervised linear-boundary margin
$m^{\mathrm{lin}}_\ell(x)$, oriented so that positive values indicate the
safe-like side and negative values indicate the unsafe-like side. For a
given evaluation cell, let $S$ and $U$ denote the safe and unsafe test
examples. At each layer $\ell$, we define the standardized separation of
the margin level as
\[
\mathrm{Sep}^{\mathrm{level}}_\ell
=
\frac{
\left|
\mathbb{E}_{x \in S}[m^{\mathrm{lin}}_\ell(x)]
-
\mathbb{E}_{x \in U}[m^{\mathrm{lin}}_\ell(x)]
\right|
}{
\sqrt{
\frac{1}{2}
\left(
\mathrm{Var}_{x \in S}[m^{\mathrm{lin}}_\ell(x)]
+
\mathrm{Var}_{x \in U}[m^{\mathrm{lin}}_\ell(x)]
\right)
}
+\epsilon
}.
\]
For the drift diagnostic, we first compute the adjacent-layer difference
\[
\Delta m^{\mathrm{lin}}_\ell(x)
=
m^{\mathrm{lin}}_{\ell+1}(x)-m^{\mathrm{lin}}_\ell(x),
\]
and apply the same standardized separation statistic:
\[
\mathrm{Sep}^{\mathrm{drift}}_\ell
=
\frac{
\left|
\mathbb{E}_{x \in S}[\Delta m^{\mathrm{lin}}_\ell(x)]
-
\mathbb{E}_{x \in U}[\Delta m^{\mathrm{lin}}_\ell(x)]
\right|
}{
\sqrt{
\frac{1}{2}
\left(
\mathrm{Var}_{x \in S}[\Delta m^{\mathrm{lin}}_\ell(x)]
+
\mathrm{Var}_{x \in U}[\Delta m^{\mathrm{lin}}_\ell(x)]
\right)
}
+\epsilon
}.
\]
We report late-layer averages over
$L_{\mathrm{late}}=\{\ell:\ell>\lfloor 2L/3\rfloor\}$, matching the
late-layer feature definitions in Section~\ref{sec:method:features}.
For drift, the average uses adjacent-layer differences whose starting
index lies in the valid late-layer range.

\begin{table}[h]
\centering
\caption{Standardized class separation for supervised linear-boundary
margin level and drift. Values are mean $\pm$ standard deviation across
evaluation cells. Each cell corresponds to a model and seed for full-test
and hard-subset evaluations, and to a model and held-out benchmark for
hard LOBO. Ratio denotes
$\mathrm{Sep}^{\mathrm{drift}}_{\mathrm{late}} /
\mathrm{Sep}^{\mathrm{level}}_{\mathrm{late}}$.}
\label{tab:snr_diagnostic}
\small
\setlength{\tabcolsep}{5pt}
\begin{tabular}{lcccc}
\toprule
Slice & Cells & Level sep. & Drift sep. & Drift / Level \\
\midrule
Full test   & 27 & $2.175 \pm 0.233$ & $0.100 \pm 0.020$ & $4.7\%$ \\
Hard subset & 27 & $1.510 \pm 0.168$ & $0.104 \pm 0.027$ & $6.9\%$ \\
Hard LOBO   & 27 & $1.204 \pm 0.337$ & $0.136 \pm 0.046$ & $11.7\%$ \\
\bottomrule
\end{tabular}
\end{table}

Across all three regimes, level separation is larger than drift
separation in every measured cell. In the hard-LOBO slice, the
late-layer drift/level separation ratio is $11.7\%$ on average, with
median $11.4\%$ and range $5.7\%$--$24.9\%$. Thus, even under benchmark
shift, finite-difference motion carries substantially weaker
class-separation signal than the absolute position of the margin profile.

\subsection{Margin and drift visualization}
\label{app:margin_drift_visualization}

\begin{figure}[h]
\centering
\includegraphics[width=0.38\textwidth]{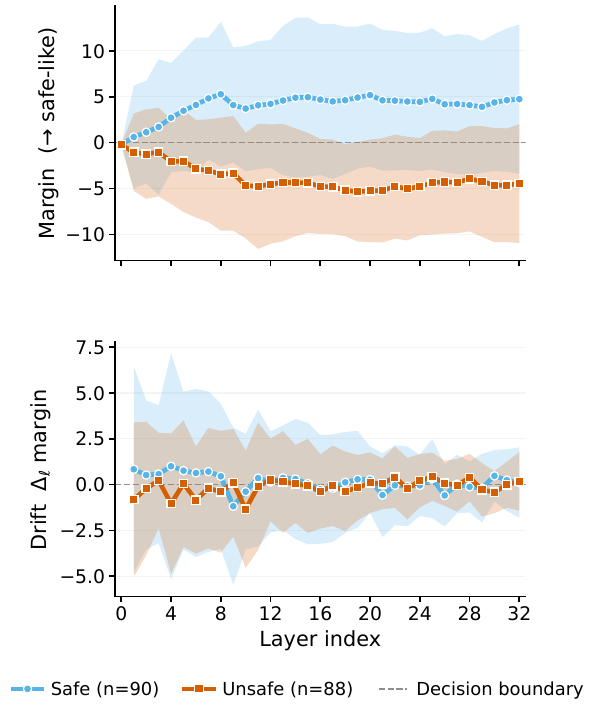}
\caption{Class-conditional linear-boundary margin and drift on
Llama-3.1-8B, BeaverTails. Lines show class means; bands show
$\pm 1$ s.d.}
\label{fig:app_margin_drift}
\end{figure}

Figure~\ref{fig:app_margin_drift} visualizes the dominant
in-distribution readout, the supervised linear-boundary margin, on
Llama-3.1-8B and BeaverTails. The margin profile separates the two
classes early and remains separated across depth: unsafe prompts move
into the negative half-plane of the per-layer boundary, while safe
prompts remain mostly positive. By contrast, the finite-difference drift
profile is much less class-separable. Both classes oscillate around
zero, and their uncertainty bands overlap substantially across layers.

\section{Drift case studies}
\label{app:drift_cases}

\subsection{Sample-level drift cases}
\label{app:drift_cases_samples}
The feature-group ablation in Section~\ref{sec:results:decomposition}
shows that layer-to-layer change contributes little to aggregate AUROC.
This does not imply that the finite-difference signal is uninformative
for every individual prompt. Rather, its effect is sparse: it can change
the decision on selected examples, but not often enough to affect pooled
metrics.

Figure~\ref{fig:drift_cases} illustrates this diagnostic role using two
ToxicChat examples where the margin-level summary and the full
Geometry-Lite feature set make different predictions. In both cases, the
margin profile alone gives an ambiguous or incorrect decision, while the
additional non-level summaries capture a late change in the profile that
moves the final prediction in the correct direction. These examples are
therefore intended as qualitative diagnostics of how drift-like features
can matter locally, not as evidence that layer-to-layer motion is a
dominant aggregate signal.

\begin{center}
\captionsetup{type=figure}
\includegraphics[width=0.82\columnwidth]{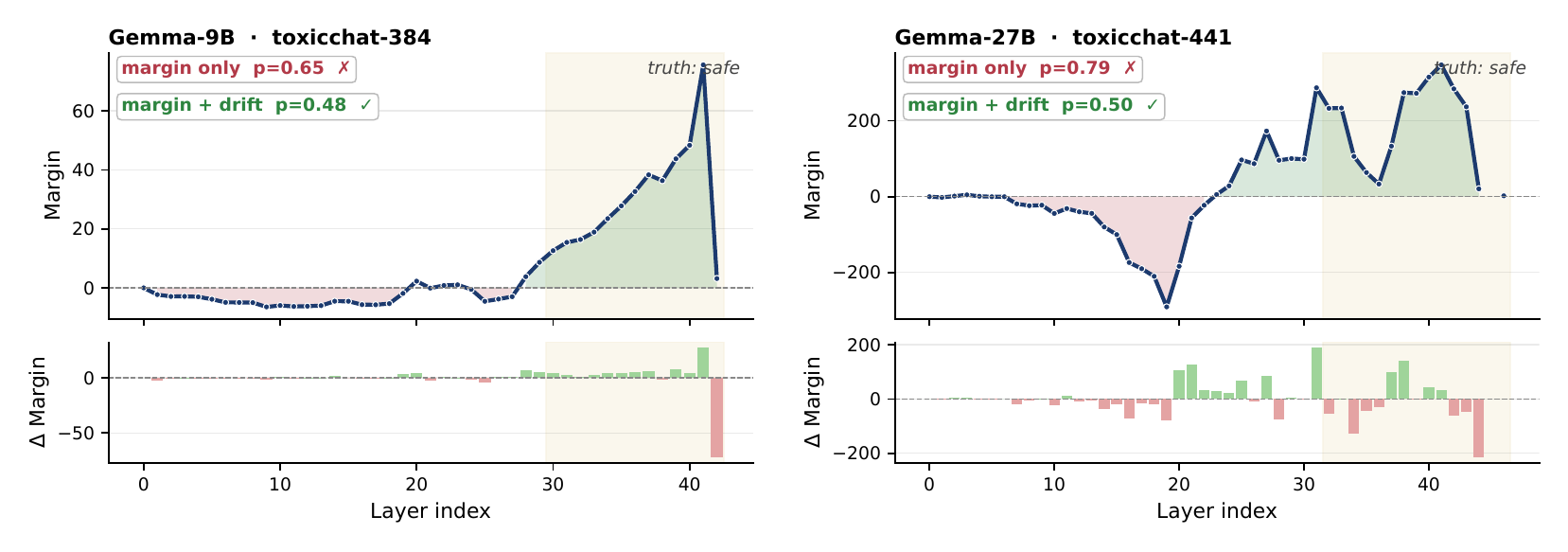}
\captionof{figure}{Illustrative late-layer correction cases on 
ToxicChat. Each example is a safe prompt misclassified by the 
magnitude-only variant; adding drift changes the decision in the 
correct direction. These cases are diagnostic rather than 
representative.}
\label{fig:drift_cases}
\end{center}

\subsection{Quantified drift analysis}
\label{app:drift_quantified}
To quantify how often layer-to-layer drift changes individual 
predictions away from the magnitude-only baseline, we count 
prediction flips at threshold $0.5$. We compare the prediction of 
\emph{magnitude\_only} ($21$ features) to the prediction of the full 
Geometry-Lite probe ($39$ features, magnitude + neg-drift + 
structural) on the same prompt across $9$ backbones and $3$ seeds.

\begin{table}[h]
\centering
\caption{Per-prompt prediction-flip counts at threshold $0.5$. 
``Drift saves'' counts cases where magnitude is wrong and full is 
correct; ``drift hurts'' counts the opposite. Net = saves $-$ hurts.}
\label{tab:drift_cases}
\small
\setlength{\tabcolsep}{6pt}
\begin{tabular}{l|cc}
\toprule
Slice & Full test & Hard subset \\
\midrule
$n$ predictions               & $23{,}481$ & $10{,}872$ \\
Magnitude only correct        & $0.8842$ & $0.7973$ \\
Full probe correct            & $0.8861$ & $0.8004$ \\
\midrule
Drift saves (count)            & $108$ & $78$ \\
Drift saves (\% of all)        & $0.46\%$ & $0.72\%$ \\
Drift saves (\% of mag misses) & $3.97\%$ & $3.54\%$ \\
\midrule
Drift hurts (count)            & $62$ & $44$ \\
Drift hurts (\% of all)        & $0.26\%$ & $0.40\%$ \\
\midrule
Net (saves $-$ hurts)          & $+46$ ($+0.20\%$) & $+34$ ($+0.31\%$) \\
\bottomrule
\end{tabular}
\end{table}

The flip rate is small in both regimes. Drift- and structural-feature 
additions recover roughly $4\%$ of the cases that magnitude-only 
misses, while introducing opposing errors at about half that rate. 
The net swing is well below $1\%$ in both regimes, consistent with 
the AUROC gap of $0.001$--$0.002$ in 
Table~\ref{tab:feat_ablation}.



\end{document}